\documentclass[format=acmsmall,screen,authorversion=true, nonacm=true]{acmart}
\usepackage[capitalise,noabbrev]{cleveref}
\usepackage{multirow}
\usepackage{subfig}
\usepackage{paralist}
\usepackage{amsmath}
\usepackage{amsfonts}
\usepackage{xcolor}
\usepackage{graphicx}
\usepackage{enumitem}
\usepackage{wrapfig}
\usepackage[linguistics]{forest}

\AtBeginDocument{%
  \providecommand\BibTeX{{%
    \normalfont B\kern-0.5em{\scshape i\kern-0.25em b}\kern-0.8em\TeX}}}

\acmJournal{CSUR}

\begin{document}

\title{A Review of the Role of Causality in Developing Trustworthy AI Systems}

\author{Niloy Ganguly}
\authornote{All authors contributed equally, names are randomly ordered}
\authornote{Presently in Indian Insitute of Technology, Kharagpur}
\email{ganguly@l3s.de}
\orcid{0000-0002-3967-186X}
\author{Dren Fazlija}
\authornotemark[1]
\email{dren.fazlija@l3s.de}
\orcid{0000-0002-9133-3526}
\author{Maryam Badar}
\authornotemark[1]
\email{badar@l3s.de }
\orcid{0000-0001-7896-0413}
\author{Marco Fisichella}
\authornotemark[1]
\email{mfisichella@l3s.de}
\orcid{0000-0002-6894-1101}
\author{Sandipan Sikdar}
\authornotemark[1]
\email{sandipan.sikdar@l3s.de }
\orcid{0000-0001-8957-7364}
\author{Johanna Schrader}
\authornotemark[1]
\email{schrader@l3s.de }
\orcid{0000-0001-9473-4635}
\author{Jonas Wallat}
\authornotemark[1]
\email{wallat@l3s.de }
\orcid{0000-0003-1239-2067}
\author{Koustav Rudra}
\authornotemark[1]
\authornote{Presently in Indian Insitute of Technology, Indian School of Mines, Dhanbad}
\email{rudra@l3s.de}
\orcid{0000-0002-2486-7608}
\author{Manolis Koubarakis}
\authornotemark[1]
\authornote{Presently in National and Kapodistrian University of Athens}
\email{koubarakis@l3s.de}
\orcid{0000-0002-1954-8338}
\author{Gourab K. Patro}
\authornotemark[1]
\email{patro@l3s.de}
\orcid{0000-0002-2435-6859}
\author{Wadhah Zai El Amri}
\authornotemark[1]
\email{wadhah.zai@l3s.de}
\orcid{0000-0002-0238-4437}
\author{Wolfgang Nejdl}
\authornotemark[1]
\email{nejdl@l3s.de }
\orcid{0000-0003-3374-2193}
\affiliation{%
  \institution{L3S Research Center, Leibniz University of Hannover}
  \streetaddress{Welfengarten 1}
  \city{Hannover}
  \state{Niedersachsen}
  \country{Germany}
  \postcode{30167}
}

\renewcommand{\shortauthors}{Ganguly and Fazlija, et al.}

\begin{abstract}
State-of-the-art AI models largely lack an understanding of the cause-effect relationship that governs human understanding of the real world. Consequently, these models do not generalize to unseen data, often produce unfair results, and are difficult to interpret. This has led to efforts to improve the trustworthiness aspects of AI models. Recently, causal modeling and inference methods have emerged as powerful tools. This review aims to provide the reader with an overview of causal methods that have been developed to improve the trustworthiness of AI models. We hope that our contribution will motivate future research on causality-based solutions for trustworthy AI.
\end{abstract}

\ccsdesc[500]{Information systems}
\ccsdesc[500]{Computing methodologies~Artificial intelligence}
\keywords{Causality, Counterfactual, Interpretability, Explainability, Robustness, Bias, Discrimination, Fairness, Privacy, Safety, Healthcare}

\maketitle
\section{Introduction}
\label{sec:intro}
The Deep Learning based systems, in recent years, have produced superior results on a wide array of tasks; 
however, they generally have limited understanding of the relationship between causes and effects in their domain~\cite{pearl2019seven}. 
As a result, they are often brittle and unable to adapt to new domains, can treat individuals or subgroups unfairly, and have limited ability to explain their actions or recommendations ~\cite{pearl2019seven,Schoelkopf_2021} reducing the trust of human users \cite{DBLP:journals/csur/KaurURD23}.
Following this, a new area of research, \emph{trustworthy AI}, has recently received much attention from several policymakers and other regulatory organizations. The resulting guidelines (e.g., 
\cite{EUEthicsGuidelines,us-gao-accountability,oecd-trustworthy}), introduced to increase trust in AI systems, make developing trustworthy AI not only a technical (research) and social endeavor but also an organizational and (legal) obligational requirement.

In this paper, we set out to demonstrate, through an extensive survey, that \emph{causal modeling and reasoning} is an emerging and  very useful tool for enabling current AI systems to become trustworthy. 
\textit{{Causality}} is the science of reasoning about causes and effects. Cause-and-effect relationships are central to how we make sense of the world around us, how we act upon it, and how we respond to changes in our environment. 
In AI, research in causality was pioneered by the Turing award winner Judea Pearl long back in his 1995 seminal paper~\cite{pearl-biometrika}. 
Since then, many researchers have contributed to the development of a solid mathematical basis for causality; see, for example, the books~\cite{pearl2009models,peters-janzing-schoelkopf-book2017,works-of-pearl}, the survey~\cite{DBLP:journals/csur/GuoCLH020} and seminal papers \cite{pearl2019seven,Schoelkopf_2021}.

The TAILOR project~\cite{tailor-srir2022}, an initiative of EU Horizon 2020, with the main objective of integrating learning, reasoning, and optimization in next-generation AI systems, in its first strategic research and innovation roadmap, identifies the following dimensions of AI systems which contribute to trustworthiness: interpretability or explainability, safety and robustness, fairness, equity and justice, accountability and reproducibility, privacy, and sustainability. This is corroborated by several recent surveys and reports~\cite{DBLP:journals/csur/KaurURD23,DBLP:journals/cacm/Wing21,DBLP:journals/ijhci/Shneiderman20,DBLP:journals/corr/abs-2004-07213,DBLP:conf/fat/JacoviMMG21}.
Following the above-mentioned works, we select a number of properties desired for the trustworthiness of AI and survey the role of causality in achieving these properties:
\begin{itemize}[leftmargin=3.5mm]
\item \textit{\textbf{Interpretability }} 
Can the AI system's output be justified with an explanation that is meaningful to the user?
\item \textit{\textbf{Fairness:}} How can we ensure that the AI system is not biased, does not discriminate (e.g., against minorities, disabled people, people of a certain gender, etc.), but provides fair prediction and recommendation?
\item \textit{\textbf{Robustness:}} How sensitive is the AI system’s output to changes in the input? Does the performance vary significantly in the case of distributional shifts?
\item \textit{\textbf{Privacy:}} Is the AI system susceptible to attacks by adversaries? Does the system leak private information? Does the AI system respect user privacy? How can we develop AI systems that ensure user privacy?
\item \textit{\textbf{Safety and Accountability:}} What are the risks for humans after an AI system is deployed in the real world? Is the AI system safe? Who or what is responsible for the actions (or failures) of the AI system after deployment? How to audit an AI system and its impact? 
\end{itemize}

There have been several current surveys on trustworthy AI~\cite{DBLP:journals/csur/KaurURD23,DBLP:journals/cacm/Wing21,DBLP:journals/ijhci/Shneiderman20,DBLP:journals/corr/abs-2004-07213,DBLP:journals/electronicmarkets/ThiebesLS21} but none has paid significant attention to the role of causality in developing trustworthy AI systems.
We provide a detailed comparison of our survey to other related surveys in \cref{sec:related_work}.
We survey the contributions of causality in upholding various trustworthy AI aspects (interpretability, fairness, robustness, privacy, safety, and auditing) in \cref{sec:interpretability,sec:fairness,sec:robustness,sec:privacy,sec:safety}.
Considering the recent interest in AI for healthcare, we discuss various trustworthy aspects needed in the health domain and survey how causality has played an important role in ensuring trust in AI-based healthcare applications (\cref{sec:health-care}). 
We also list available datasets, tools and packages relevant to causality-based trustworthy AI research and development (\cref{sec:appendix_interpretability,sec:appendix_fairness,sec:appendix_robustness,sec:appendix_privacy,sec:appendix_safety,sec:appendix_healthcare}).
\section{Preliminaries on Causality}
\label{sec:causality}
The  two most powerful  causal frameworks are   \emph{(i) structural causal models}~\cite{pearl-biometrika,pearl2009models} developed in AI, and Rubin's \emph{(ii) potential outcomes framework}~\cite{potential-outcomes-rubin} developed in Statistics. Here, we give short introductions to structural causal models (\cref{sec:scm}), the potential outcomes framework (\cref{sec:po}), and then discuss how they are connected \cref{sec:scm_po_connection}. \emph{Causal graphical models}~\cite{DBLP:books/daglib/0023012} is another popular framework that we do not cover here.
Before discussing the frameworks, we first list some basic notations, and discuss some fundamental concepts on causality: the difference between correlation and causation in \cref{sec:correlation_causation}, and the ladder of causation through association, intervention, and counterfactuals in \cref{sec:ladder_of_causation}.
~\\\textbf{Notations:}
We use non-boldface uppercase letters (e.g., ${X}$) to denote single random variables and non-boldface lowercase letters (e.g., ${x}$) to denote their values. Boldface uppercase letters (e.g., $\mathbf{X}$) denote a collection of variables, and boldface lowercase letters (e.g., $\mathbf{x}$) their values. Random variables can have continuous or discrete or categorical values. 
 
\subsection{Correlation vs. Causation}
\label{sec:correlation_causation}
Statisticians and others often state that ``correlation does not imply causation'' and illustrate it with anecdotal examples. 
\citet{Schoelkopf_2022}, for example, cite the correlation between chocolate consumption and the number of Nobel prizes per capita from~\cite{chocholate-example}, which obviously should not lead us to think that increasing chocolate consumption helps in winning a Nobel prize.
We first define causal effect relationships and causal graphs before formally discussing correlation vs. causation.
~\\{\bf Causal effect relationships:} 
Following the definition given by~\citet{Schoelkopf_2022}, we say that a random variable $X$ has a \emph{causal effect} on a random variable $Y$ if there
exist $x \ne x'$ such that the probability distribution of $Y$ after intervening on $X$ and setting it to $x$ differs from the probability distribution of $Y$ after setting $X$ to $x'$. This notion of a causal effect is \emph{unidirectional} and \emph{asymmetric}; this is not true for correlation which is a symmetric relationship. 
Note Judea Pearl~\cite{pearl-biometrika,pearl2009models} first presented the approach of \emph{intervention} for finding causal effects.
Causal effect relationships are usually represented graphically using causal graphs.

\begin{wrapfigure}{l}{0.35\textwidth}
\includegraphics[width=0.9\linewidth]{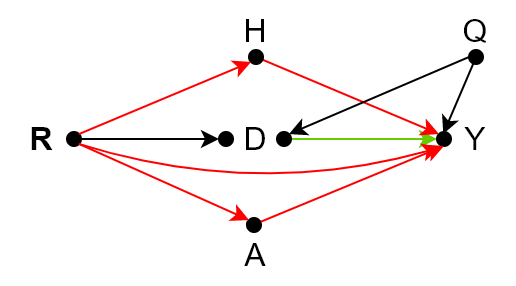} 
\caption{Causal graph of a university admission dataset representing fair and unfair causal pathways: R, D, Q, H, A, Y are causal variables representing  ``race'', ``choice of department'', ``qualification'', ``hobbies'', ``address'', and ``admission outcome'' respectively.}
\label{fig:fair and unfair}
\end{wrapfigure}

\noindent{\bf Causal graphs: \label{causal_graph}} A \emph{causal graph} is  a directed acyclic graph where nodes are the variables/attributes of the data under consideration and edges denote \emph{direct} causal effects. For example, consider Figure \ref{fig:fair and unfair}, which shows a causal graph related to a dataset of university admissions. In this graph, $R, D, Q, H, A$, and $Y$ are causal variables representing ``race'', ``choice of department'', ``qualification'', ``hobbies'', ``address'', and ``admission outcome''. In a causal graph, a \emph{causal path} is an acyclic  sequence of adjacent nodes from the starting node to the last node. For example, in Figure \ref{fig:fair and unfair}, $R \rightarrow D \rightarrow Y$ and $R \rightarrow H \rightarrow Y$ are examples of causal paths. In this figure, the variables $R, H, D, Q$, and $A$ directly affect the outcome variable $Y$, whereas the variable $R$ also indirectly causally affects $Y$ through $H, Q,$ and $A$. 

The connection between correlation and causation has been expressed by~\citet{Reichenbach_1956}
using the \emph{common cause principle} (taken here verbatim from~\cite{Schoelkopf_2022}): \textit{If two random variables $X$ and $Y$ are statistically dependent, then there
exists a random variable $Z$ which causally influences both of them and which explains all their dependence in the sense of rendering them conditionally independent. As a special case, $Z$ may coincide with $X$ or $Y$}.
The three possible cases for random variables $X, Y$ and $Z$ are: 
(i) random variables $X$ and $Z$ coincide,
(ii) random variables $Y$ and $Z$ coincide, or
(iii) $Z$ is a new unobserved variable which causally influences both $X$ and $Y$.
For the example of chocolate consumption (variable $X$) vs. the number of Nobel prize winners (variable $Y$), obviously, $Y$ is not a causal effect of $X$ and vice versa. The common cause principle then tells us that there is another random variable $Z$ (e.g., economic prosperity) that has $X$ and $Y$ as causal effects. 
In this example, the unobserved variable $Z$ introduces a \textit{spurious correlation} (i.e., a correlation that does not stem from a direct causal effect relationship between two observed variables). Such variables are referred to as \textit{confounders} or \textit{confounding variables}.

\subsection{\bf The Ladder of Causation} 
\label{sec:ladder_of_causation}
\citet{pearl-why} presented the \emph{ladder of causation} as an informal way of explaining the principles of causality. 
The ladder of causation has three levels:
\emph{seeing, doing, and imagining}, and forms a directional 3-level hierarchy in the sense that questions at a level can only be answered if information from previous level(s) is available~\cite{pearl2019seven}.
Next, we discuss them in detail.
~\\\textbf{\textit{(1) Association:}} The first level (seeing or observing) corresponds to detecting \emph{associations} in the domain under study and making predictions based on these associations. Most animals, as well as most of today's (purely statistical) machine learning systems (including sophisticated deep learning systems), are on this level. The same is true for popular probabilistic frameworks such as Bayesian networks since this level is characterized by conditional probability sentences e.g., the sentence $P(Y=y|X=x) = 0.8$.
For some associations, it might be easy to find causal interpretations, while for others it might not. \textit{But agents on this level of the ladder, cannot differentiate between a cause and an effect.} Therefore, they cannot answer ``why'' questions.
~\\\textbf{\textit{(2) Intervention:}} The second level (doing) corresponds to \emph{interventions}. Most tool users are on this level if they plan their actions and not imitating the actions of others. To see the effects of interventions, we can do experiments (e.g., a randomized controlled trial).  Pearl and his colleagues have formalized reasoning on this level of the ladder of causation using the \emph{do calculus}~\cite{pearl-why,pearl2009models,works-of-pearl}.
This level of the ladder is characterized by expressions such as $P(Y=y|do(X=x),\ Z=z)$ which means ``the probability of event $Y = y$ given that we intervene and set the value of $X$ to $x$ and subsequently observe event $Z = z$''~\cite{pearl2019seven}. 
~\\\textbf{\textit{(3) Counterfactuals:}} The third level (imagining) corresponds to \emph{counterfactual reasoning} and its associated modal reasoning capabilities (e.g., retrospection, introspection, etc.). On this level, we can imagine worlds that do not exist, including worlds that contradict the world in which we live, and infer why the phenomena we have observed in our domain have taken place. This level of the ladder is characterized by expressions of the form $P(Y=y_{X=x} |X=x', Y=y')$ which stand for ``the probability that event 
$Y = y$ would be observed had $X$ been $x$, given that we actually observed the events $X=x'$ and $Y=y'$''~\cite{pearl2019seven}. Such expressions can be computed only if we have a causal formalism such as the structural causal networks presented in Section~\ref{sec:scm} below. Given such a structure, one can then also formalize level 3 reasoning through do calculus.

\subsection{\bf Structural Causal Models (SCM)}
\label{sec:scm}
Structural causal models consist of causal graphs and structural equations.
Following the definition by~\citet{bareinboim2022}, a \emph{structural causal model (SCM)} $\mathcal{M}$ is a 4-tuple
$\langle \mathbf{U}, \mathbf{V}, \mathcal{F}, P(\mathbf{U}) \rangle$
where:
$\mathbf{U}$ is a set of background variables, also called \emph{exogenous} variables, that are determined by factors outside the model;
$\mathbf{V}$ is a set of variables, called \emph{endogenous} variables, that are determined by other variables in the model i.e., variables in $\mathbf{U} \cup \mathbf{V}$;
$\mathcal{F}$ is a set of functions $\{ f_1, f_2, \ldots, f_n \}$ such that each $f_i$ is a mapping from the respective domains of $\{ U_i \} \cup \mathbf{PA}_i$ to $V_i$, where $U_i \in \mathbf{U}$, the variables $\mathbf{PA}_i$ are the \emph{parents} of variable $V_i$ and are such that $\mathbf{PA}_i \subseteq \mathbf{V} \setminus \{ V_i \}$, and the entire set $\mathcal{F}$ forms a mapping from $\mathbf{U}$ to $\mathbf{V}$, so, for $i = 1, \ldots, n$, each $f_i \in \mathcal{F}$ is such that
$V_i  \leftarrow f_i(\mathbf{PA}_i, U_i)$,
i.e., it assigns a value to $V_i$ that depends on the values of the parents of $V_i$ and the value of the exogeneous variable $U_i$;
$P(\mathbf{U})$ is a probability function defined over the domain of $\mathbf{U}$.
The exogenous ones are determined ``outside'' of the causal model (therefore assumed to be independent), and their associated probability distribution $P(\mathbf{U})$ gives a summary of the state of the world
outside the domain. 

The expression $V_i  \leftarrow f_i(\mathbf{PA}_i, U_i)$ in the above definition are called \emph{structural equations}\footnote{We follow~\cite{bareinboim2022} and avoid using an equality sign for structural equations since their interpretation is that of an assignment statement; they should not be interpreted as algebraic equations that can be solved for any variable.}.
The structural equations \emph{induce a causal graph} for the SCM. The nodes of the graph are the variables in the set $\mathbf{V}$, and there is a directed edge from each variable in $\mathbf{PA}_i$ to $V_i$ for all $i$. In other words, the structural equations encode the direct causal effects for each edge in the causal graph, and they can be used to
determine the value of each endogenous variable $V_i$ in terms of the values of the exogenous variable $U_i$ and the values of the endogenous variables $\mathbf{PA}_i$ that are parents of
$V_i$. The causal processes encoded by structural equations are assumed to be invariant unless we explicitly intervene on them using the do calculus of the second level of the ladder of causality~\cite{bareinboim2022}.

SCMs can be used for various inference tasks that the forthcoming sections of this paper show to be useful for achieving the trustworthiness of AI systems. 
Some well-known tasks are listed below.
\begin{itemize}[leftmargin=3.5mm]
    \item The first such task is \textbf{causal reasoning}, which is the process of deriving conclusions from a causal model e.g., an SCM. SCMs can also be used to study the effects of interventions or distribution changes or carry out counterfactual reasoning~\cite{pearl2009models,peters-janzing-schoelkopf-book2017}. 
    \item The opposite task of causal reasoning is \textbf{causal discovery} or \textbf{causal structure learning}, which is the process of inferring SCMs from data, assumptions, empirical observations, or from data under interventions or distribution changes~\cite{pearl2009models,peters-janzing-schoelkopf-book2017}.
    \item Another interesting task is \textbf{causal mediation}, which is the process of looking for the mechanism that explains how a cause is connected with an effect~\cite{DBLP:conf/uai/Pearl01}. In causal mediation, we may have an SCM $X \rightarrow Z \rightarrow Y$ where $X$ is the cause, $Y$ is the effect, and $Z$ is the \emph{mediator} i.e., a variable that can be used to answer the question ``Why $X$ causes $Y$?''. For example, the SCM $Citrus\ Fruit \rightarrow Vitamin\ C \rightarrow Scurvy$, explains why citrus fruits were important in helping sailors avoid scurvy in the 1800s~\cite{pearl-why}. In such networks, $Y$ is an \emph{indirect} effect of $X$ as opposed to a \emph{direct} effect, which would have been denoted by $X \rightarrow Y$.
\end{itemize}

\subsection{\bf Potential Outcomes (PO) Framework}
\label{sec:po}
The \emph{potential outcomes} approach to causality was developed to make statistically valid statements even in cases where randomized controlled studies are not or only partially possible. 
This section gives a brief introduction to the PO framework~\cite{potential-outcomes-rubin}. 

\textbf{The missing data problem in individualized treatment effect (ITE):} We quote an example from~\citet{ozer2022}.
Consider two possible outcomes for one patient having resectable gallbladder cancer, ``survival'' and ``no survival''. 
The task is to measure the causal effect of a treatment (e.g., chemotherapy).
Here, we have a binary \emph{treatment} variable $T$ with $T=1$ if the person is getting the chemotherapy and $T=0$ otherwise (\emph{control}), and its effect on an \emph{outcome} variable $Y$ (typically a measure of health) is to be found.
Note that $Y_i(1)$ and $Y_i(0)$ are unobserved outcomes for $T=1$ and $T=0$ respectively.
Then the \textit{individualized treatment effect} can be captured by the difference between the two potential outcomes, i.e., $\tau_i=Y_i(1)-Y_i(0)$.
Since only one of the outcomes will take place, we can not have both $Y_i(1)$ and $Y_i(0)$ for an individual.
To overcome this problem, the PO framework takes some assumptions and estimates average causal effects over a population instead of individual effects, which we detail next.

\textbf{Assumptions to overcome the missing data problem:} 

The PO framework relies on the following assumptions: (a) \textbf{The stable unit treatment value assumption.} The observation on one unit should be unaffected by the assignment of treatments to the other units; it is a reasonable assumption in many situations like controlled studies, since different units can form independent samples from a population. 
(b) \textbf{The consistency assumption.} The potential outcome for an individual remains consistent and converges. Any variation within the exposure group (treatment or control) would result in the same outcome for that individual.
Using these assumptions, the PO framework estimates average causal effects over a population, as detailed next.

\textbf{The average treatment effect (ATE) in PO:} 
Since the ITE $\tau_i$ ($Y_i(1) - Y_i(0)$) under a deterministic treatment model can not be found, an \textit{average treatment effect} on a population is calculated. 
The difference between the expected outcome in the treatment group and the expected outcome in the control group is expressed as  $\mathbb{E}[Y|1]-\mathbb{E}[Y|0]$. 
So, the ATE can be calculated by first randomizing the treatment and control group assignments of units, followed by a comparison of mean outcomes for treated and untreated units.
In real-world settings, some covariates (e.g., smoking and drinking habits) also affect the outcome along with the treatment.
Thus, they must be taken into account. 

\textbf{Propensity score matching to take care of covariates:} 
In cases of covariates, one must balance the covariate distribution between the treated and untreated cohorts, {\it propensity score matching}~\cite{rosenbaum1983} is one of the most popular methods used for this. 
Propensity scores are probabilities of units being assigned to different treatment groups based on the observed covariates, and these are estimated using logistic regression over the covariates.
Now the method basically aims to make two groups comparable to each other in terms of covariates, thereby accounting for selection bias~\cite{cortes2008sample}. 
Essentially, each patient from one group is matched with another one for the other group based on the propensity scores. 
Units (patients) with no matching from the other treatment group are often removed and not used in the ATE calculation. 
Other measures, such as the \textit{Mahalanobis} metric, were excluded from this review. We refer to~\cite{stuart2010matching} for an extensive overview of matching methods, including definitions of prominent matching metrics.

\subsection{\bf Connection between SCM and PO}\label{sec:scm_po_connection} The above discussion has assumed that individuals (or units) are binary quantities. However, this is not a good assumption when dealing with complex units such as people e.g., in the health domain. In these cases, potential outcomes can be defined as random variables and a clear connection to the structural equations with exogenous noise variables in SCMs can be defined. This observation results in the equivalence of the two frameworks, a result which has been originally shown in~\cite{pearl2009models}.~\cite{Schoelkopf_2022} explains this equivalence of the two frameworks in the following simple way:
\[ Y_i(t) = Y\ |\ do(T = t) \mbox{ in an SCM with } \mathbf{U} = \mathbf{u}_i \]
This informally means that an individual $i$ in the PO framework corresponds to a particular value of 
an exogenous variable $U_i$ in an SCM. The potential outcome is
deterministic once we know $\mathbf{U}$, but since we do not observe $\mathbf{u}_i$, the counterfactual outcome is treated as a random variable~\cite{Schoelkopf_2022}.
\subsection{Prior Surveys}
\label{sec:related_work}
Some recent survey papers have discussed causality and trustworthy machine learning. Most of them focus either exclusively on causality~\cite{gao2022causal,scholkopf2022causality} or trustworthy aspects of machine learning~\cite{kaur2022modeling}, only a few papers tried to cover both causality and different aspects of trustworthiness~\cite{kaddour_2022_causalmlsurvey,Schoelkopf_2021}.

Apart from general coverage of important trustworthiness aspects, some of the surveys explicitly covered interpretability and fairness aspects in detail. 
~\citet{guidotti2018survey} and \citet{linardatos2020explainable} provided a review of machine learning interpretability methods that deal with explaining black box models.~\citet{zhou2021evaluating} performed a survey on evaluation metrics of explainability.  
\citet{makhlouf2020survey} pointed out the difficulty in choosing a particular fairness notion in a given domain and scenario. \citet{mehrabi2021survey} conducted a survey focused on the application of AI for fairness-aware learning in various domains. \citet{moraffah2020causal} performed a survey on causal inference on model, example-based interpretability, and fairness. In contrast to the above approaches, we provide detailed coverage of causality-based methods over a wide variety of trustworthy aspects.

\citet{kaddour_2022_causalmlsurvey} and \citet{Schoelkopf_2021} provide an overview of the application of causality to address trustworthy aspects. \citet{kaddour_2022_causalmlsurvey} discussed different causal approaches, such as causal supervised learning and causal generative modeling, and their applications on explainability, fairness, and robustness. 
They did not cover the application phases of different methods in detail, i.e., whether causal approaches could be applied in pre-processing, in-processing, or post-processing stages, and did not provide more extensive coverage on different aspects of trustworthiness, though.  Our work complements these prior works and extends the discussion to robustness, privacy, safety, and accountability. Apart from structural causal models, we also include a discussion on the PO framework which helps in causal analysis from a statistical perspective.

Table~\ref{table:causality_survey_summary} provides the coverage statistics of the existing surveys. Note that, this table covers the generic and domain-specific surveys that used causality as a driving force to achieve trustworthiness. It is evident from the table that our survey covers the causality and trustworthiness aspects in more detail.

In the next sections (\cref{sec:interpretability,sec:fairness,sec:robustness,sec:privacy,sec:safety}), we discuss the listed trustworthy AI properties one by one and discuss how causality is helpful in upholding them in the context of AI.

\begin{table}[tb]
\small
\center
\caption{{\bf Comparison of our survey with related causality and trustworthy-based survey papers.}}
\resizebox{\textwidth}{!}{
\begin{tabular}{|l||c|c|c|c|c|c|c|c|}
\hline
{\bf Surveys} & \multicolumn{7}{c|}{\bf Trustworthiness Aspects }  & {\bf Domain} \\ \cline{2-8}
 & {\bf Interpretability} & {\bf Fairness} & {\bf Accountability} & {\bf Robustness} & {\bf Privacy} & {\bf Safety} & {\bf Security} & \\
\hline
 \citet{kaddour_2022_causalmlsurvey} & X & X & - & X & - & - & -  & -\\
\hline
\citet{Schoelkopf_2022} & - & - & - & X & - & - & -  & -\\
\hline
\citet{guidotti2018survey} & X & - & - & - & - & - & - &  -\\
\hline
\citet{linardatos2020explainable} & X & - & - & - & - & - & - &  -\\
\hline
\citet{zhou2021evaluating} & X & - & - & - & - & - & - &  -\\
\hline
\citet{makhlouf2020survey} & - & X & - & - & - & - & - &  -\\
\hline
\citet{mehrabi2021survey} & - & X & - & - & - & - & - &  -\\
\hline
\citet{moraffah2020causal} & X & X & - & - & - & - & - & -\\
\hline
\citet{zhang2021} & X & X & - & X & - & - & - &  Healthcare\\
\hline
\citet{Sanchez2022} & X & X & - & X & - & - & - &  Healthcare\\
\hline
\citet{vlontzos2022} & X & X & - & X & - & X & - &  Healthcare \& \\
 &  &  &  &  &  &  &  & Image analysis\\
\hline
\citet{gao2022causal}& - & - & - & - & - & - & - &  Recommendation\\
\hline
Our Survey & X & X & X & X & X & X & X  & -\\
\hline
\end{tabular}}
\label{table:causality_survey_summary}
\end{table}

\section{Causality and Interpretability}
\label{sec:interpretability}
Understanding latent models is one of the central challenges in the development and deployment of machine learning applications. 
Recent trends have shown that interpretability is sacrificed for overparameterization and the promised generalizability~\citep{gpt-3,ramesh:2022:arxiv:dalle2}. 
However, to apply these models in high-stakes scenarios such as the legal or medical domain, we will need (and potentially be legally required by the EU AI Act\footnote{Article 13, https://eur-lex.europa.eu/legal-content/EN/TXT/?uri=CELEX\%3A52021PC0206})
to build understandable models. To ensure that the model's explanation communicates the true reasons behind a prediction, the explanation itself should preferably be causal.
This section aims to provide an overview of the early stages of causality in existing interpretability research, its promises, and how it might help build more trustworthy models.

\subsection{Preliminaries}
\textit{Evaluation Criteria.} 
Faithfulness and causability are two important criteria to measure the interpretability of non-causal methods. 
Faithfulness measures how well the explanation matches the actual underlying processes of the model. To gain human confidence, the accuracy of the explanation is of immense importance in high-stake scenarios, such as medical or health applications. Various metrics have been proposed to measure faithfulness, e.g., feature attribution methods.
Causability measures how well explanations depict the causal structure of the problem and can be assessed through a user study using the System Causability Scale (SCS), which uses a Likert scale and is similar to the well-established System Usability Scale (SUS)~\cite{Holzinger2020}. 
A causable explanation helps the recipient build a correct mental model of the problem, which has been shown to greatly impact the user's trust in the system~\citep{shin2021causability}. For an effective AI system, the target audience must be taken into account when evaluating the causability of the method, that is, causal explanations of a model used in the medical domain should provide more details when dedicated to medical experts than an explanation presented to patients.

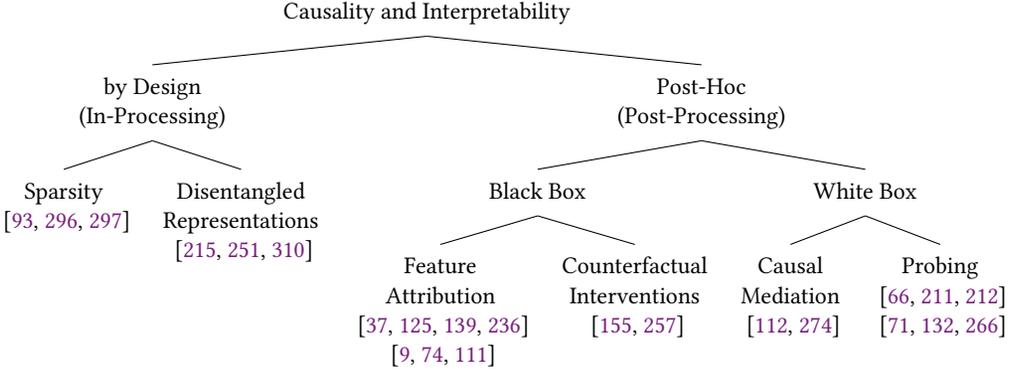
\begin{figure} 
\small
  \centering   
    \begin{forest}
      for tree={
        parent anchor=south,
        child anchor=north,
        fit=band,
      }
      [Causality and Interpretability
        [by Design \\ (In-Processing)
          [Sparsity \\ ~\cite{guyon2007causal,Yu2020,Yu2021}] 
          [Disentangled \\ Representations \\~\cite{zheng:2021:www:disentangeluserinterestConformity,si:2022:www:causalRecSearch,reddy:2022:AAAI:candle_disentangled_dataset}]
        ]
        [Post-Hoc \\ (Post-Processing)
          [Black Box
            [Feature \\ Attribution \\~\cite{Kim_2017_ICCV,Schwab2019,cao:2022:acl:promtLMBiasCausal,li:2022:acl:PTCaptureFactsCausal} \\~\cite{Janzing2020,NEURIPS2020_0d770c49,alvarez2017causal}
            ]
            [Counterfactual \\ Interventions \\~\cite{Tan2021,Mahajan2019}
            ]
          ]
          [White Box
            [Causal \\ Mediation \\~\cite{DBLP:conf/nips/VigGBQNSS20, jeoung:2022:arxiv:DebiasCausalMediation}]
            [Probing \\~\cite{DBLP:journals/tacl/ElazarRJG21, ravfogel:2020:acl:INLP,ravfogel:2022:acl:RLACE} \\~\cite{tucker:2021:IJCNLP:modifiedthat,DBLP:conf/acl/LasriPLPC22,feder:2021:coli:CausaLM}]
            ]
        ]
      ]
    \end{forest}
    \caption{Structure of approaches introducing causality in interpretability.}
    \label{fig:interpretability_tree}
\end{figure}

Like traditional interpretability methods, causal approaches are either interpretable by their model design or are methods that provide post-hoc explanations for non-interpretable models. In this section, we first provide a brief overview of methods that are causally interpretable by design and then focus on causal post-hoc interpretability. Figure~\ref{fig:interpretability_tree} visualizes an overview of the different approaches and the application of causal interpretability.

\subsection{\bf Causal Interpretability By Design}
With the recent surge of interest in causality, existing approaches to interpretable ML, such as feature selection and disentanglement representations, have also been extended to include causality in interpretable methods. In the following, we will give an overview of both approaches in causality.

\subsubsection{Sparsity (Causal Feature Selection)} 
Feature selection categorizes the data’s attributes into relevant (irrelevant) features, i.e., features that do (not) provide predictive information regarding a target variable, and redundant features, i.e., multiple features with the same predictive information, and removes irrelevant and redundant features from the variable-subset of interest~\cite{guyon2007causal}.
Causality is introduced into feature selection methods to distinguish between causally relevant and only co-occurring features when choosing  the redundant features to keep as relevant attributes~\cite{guyon2007causal}, which may lead to higher accuracy. 
To introduce a causal ordering of the features, Bayesian networks (BNs) can be used in causal feature selection to represent causal relationships among features in a DAG by interpreting directed edges as cause-effect relationships. Under the faithfulness assumption, the Markov boundary (MB) of a variable of interest in the BN describes the variable’s local causal relationships, i.e., its parents, children, and spouse~\cite{Yu2020}.
In most cases, a model has only one variable of interest, i.e., the target variable, making it unnecessary to learn a full BN. Instead, these approaches~\cite{Aliferis2010a,Aliferis2010b} only learn the MB for the variable of interest, using either constraint-based methods based on conditional independence tests or score-based methods that combine greedy search with scoring functions~\cite{Yu2020}.
However, causal feature selection methods generally are less efficient than their non-causal counterparts due to conditioning on the other features and also are less reliable when the data set is small with high dimensionality~\cite{Yu2021}. 
For a more detailed overview, the reader is referred to a comprehensive review of causal feature selection by Yu et al.~\cite{Yu2020}. They also provide a collection of relevant methods in their package CausalFS.

\subsubsection{Disentangled Representations}
A perfectly disentangled latent representation where each dimension represents a human-understandable concept would naturally be interpretable. 
However, achieving a fully disentangled representation is not feasible in the general case, as approaches require specific training data~\citep{shen:2017:cvpr:residualImageDisentanglement}, supervision~\citep{nitzan:2020:acm_graph:faceIdentityDisentanglement}, or predefined concepts~\citep{liu:2022:ieee_image:disentangledfaceediting}. 
Nevertheless, there are approaches using SCMs (Section~\ref{sec:scm}) to identify confounders and then disentangle representations to produce more interpretable models. 
\citet{reddy:2022:AAAI:candle_disentangled_dataset} propose a dataset to investigate the causal effects in disentangled representations. The proposed dataset contains images of geometric shapes with varying properties (e.g., shape, background, color, size) generated according to an underlying SCM. This enables investigations to determine whether learned disentangled representations follow the same causal structure as used to generate the data. 
\citet{zheng:2021:www:disentangeluserinterestConformity} build a structural causal graph and note that the predictions of recommendation systems are using the intertwined information of user interest, as well as the general popularity of items. Therefore, they propose a multitask learning framework to causally disentangle these properties, where individual interest and popularity representations are learned on separate auxiliary datasets and later concatenated for the recommendation. In addition to the added interpretability of the causally disentangled representation, they also report increased robustness in the non-IID setting~\citep{zheng:2021:www:disentangeluserinterestConformity}. 
\citet{si:2022:www:causalRecSearch} continue this line of work by combining search and recommendation data to disentangle latent representation into a causally-relevant personalization part and a causally non-relevant part.

\subsection{Post-Hoc Causal Interpretability}
Post-hoc causal interpretation methods aim at causally explaining existing non-interpretable models after they have been trained, thus supplementing the good performance of complex models with human-understandable causal explanations. Causal methods for explaining non-interpretable models can be divided into two groups: black-box and white-box approaches. 
\subsubsection{Black-Box} Model agnostic approaches treat the model as a black-box and provide an explanation considering the input and output to make the model's predictions human-comprehensible without requiring access to the model's parameters. Approaches for causally explaining black-box models can be separated into two main categories: Feature attribution and counterfactual intervention methods. 

\noindent\textbf{Feature Attribution:} 
Feature attribution methods quantify the input feature's contribution to the prediction, but unlike non-causal approaches, causal approaches aim to retrieve only causally relevant features.
In causal filtering, counterfactual input representations are generated and evaluated by masking potentially influential features, e.g., determined by attention and measuring the deterioration in performance.

\citet{Kim_2017_ICCV} apply this method to real-time videos to estimate the features' causal influence on the prediction. \citet{Schwab2019} use the importance distribution determined by masking to build a causal explanation model in parallel with the prediction model and minimize the Kullback-Leibler divergence between their importance distributions.
Causal methodologies have also been applied in prompt-based interpretability.~\citet{cao:2022:acl:promtLMBiasCausal} investigate risk factors and confounders by prompting language models to complete a sentence to elicitate whether the model learned certain information from pre-training.
To do so, they built an SCM to identify risk factors by backdoor-paths and propose causal interventions to study the methodologically induced biases. 
\citet{li:2022:acl:PTCaptureFactsCausal} further investigates how the input sentence formulation can influence the model predictions by constructing counterfactual model inputs. To do so, they mask potentially relevant context words and study the impact on the amount of factual knowledge retained by the model. Their experiments suggest that co-occurrence, as well as the closeness of the subject (e.g. ``Albert Einstein'') with the object (e.g. the birthplace), are highly relevant.

\citet{NEURIPS2020_0d770c49} extend the notion of Shapley values to asymmetric Shapley values (ASV) and build up a causal framework to quantify the contribution of a feature to the model's prediction. 
However, this notion requires at least some knowledge about the causal ordering of features as it otherwise reduces to classical symmetric Shapley values without ordering.
\citet{Janzing2020} introduce a causal view to SHAP, considering the model's inputs as causes of the output. They argue for the use of interventional conditional distributions in SHAP to quantify the contribution of each observation to the output. This approach allows for determining the causal relevance of the input features without prior knowledge about the features' causal relationships. 
\citet{alvarez2017causal} use perturbation to explain specific input-output pairs for any input-output structured model by constructing a dense bipartite graph of perturbed inputs and their output tokens to infer a causal model using Bayesian logistic regression. A graph partitioning framework derives an explainable dependency graph by minimizing the \textit{k}-cut. The perturbations introduce uncertainty information via the frequency with which each token occurs and thus allow one to reveal flaws or biases in a model. 

\noindent\textbf{Counterfactual Interventions:} 
\citet{Tan2021} use counterfactual interventions to explain decisions of recommendation models. To create counterfactuals, the authors manipulate input items while optimizing for minimally changed items that reverse the recommendation. The minimal changes now serve as explanations for the original recommendations.
\citet{Mahajan2019} propose a causal proximity regularizer to incorporate knowledge on causal relations from an SCM and thus constrain generated counterfactual explanations to be actionable recourses that are feasible in real-world settings.

\subsubsection{White-Box} 
White-box interpretability methods, in contrast to black-box methods, require access to the model parameters and thus are also called model introspective approaches. Causal approaches use causal mediation or probing. 

\noindent\textbf{Causal Mediation:}
To identify the components of a model responsible for a biased prediction, ~\citet{DBLP:conf/nips/VigGBQNSS20, jeoung:2022:arxiv:DebiasCausalMediation} introduce the method of causal mediation analysis.  
More formally, causal mediation analysis is a method to examine the intermediate processes where independent variables affect dependent variables.
By intervening on the model and measuring direct and indirect effects,~\citet{DBLP:conf/nips/VigGBQNSS20} investigate gender bias in individual neurons and attention heads and find a small subset of attention heads and neurons in the intermediary layers, which are responsible for biased predictions. 
\citet{jeoung:2022:arxiv:DebiasCausalMediation} follow a similar causal mediation setup but use it to investigate the effects of debiasing techniques, finding that the debiased representations are robust to fine-tuning.

\noindent\textbf{Probing:}
Probing usually involves training a small classifier to predict a property of interest from the model's latent embeddings.
\citet{DBLP:journals/tacl/ElazarRJG21} construct a counterfactual embedding without the property of interest (POI). They then infer the usage of POI during inference if the performance drops after the removal of POI information.  
To construct these counterfactual embeddings, several methodologies have been proposed - either by iteratively training classifiers to identify relevant dimensions in the embedding space~\citep{ravfogel:2020:acl:INLP}, posing the removal as a minimax~\citep{ravfogel:2022:acl:RLACE}, or as a gradient-guided search problem~\citep{tucker:2021:IJCNLP:modifiedthat}.~\citet{DBLP:journals/tacl/ElazarRJG21} utilize this \textit{causal probing} approach and study the effects of part-of-speech knowledge on language modeling. 
\citet{DBLP:conf/acl/LasriPLPC22} (causally) probe language models on the usage of grammatical number information, finding that this resides in different layers depending on the token type (verbs/nouns). Furthermore,~\citet{tucker:2021:IJCNLP:modifiedthat} utilize their gradient-guided counterfactual creation method and find evidence for language models using tree distance-based embeddings to represent syntax. 
Finally,~\citet{feder:2021:coli:CausaLM} investigate the effect of concepts like the presence of political figures on model prediction. Using adversarially trained counterfactual representations (without the concept under investigation), they contrast the classification performance of the standard and counterfactual representations.

\subsection{Conclusion}
The surge in causality papers over the last few years also affected the interpretability field. We believe that either causal - and therefore interpretable by design - models or the usage of causal (post-hoc) interpretability for standard ML models will be paramount for building trustworthy models. Existing causal methods~\cite{DBLP:conf/nips/VigGBQNSS20, jeoung:2022:arxiv:DebiasCausalMediation} provide information on the parts of an ML model that causally affect predictions and thus also provide information on biased or erroneous models. However, the underlying processes that lead to biased predictions and, in turn, the best approach to counteract these biases is still not well understood, and methods need to be developed in that direction. We observe structural causal models to be generated and used for many problems but still lack standardized procedures for evaluation. 
For example, an assessment of the extent to which the provided explanations help humans to causally understand a model's output would be helpful.
The SCS~\cite{Holzinger2020} provides a suitable framework for causality evaluation, however, its effectiveness depends upon a wider adoption of user studies in evaluating AI research.
\section{Causality and Fairness}
\label{sec:fairness}

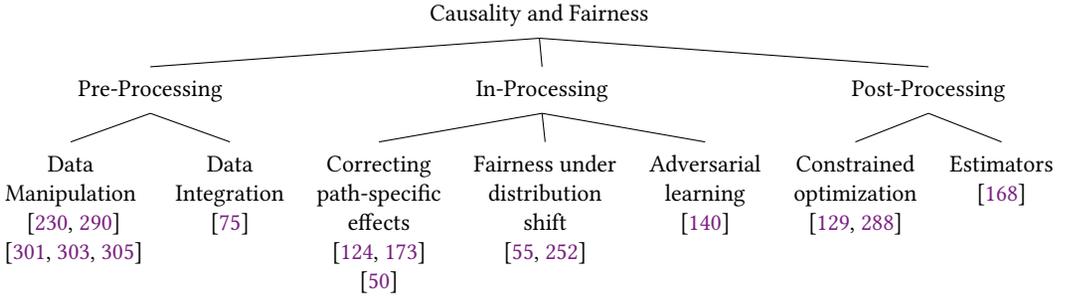
\begin{figure}
  \small
  \centering   
    \begin{forest}
      for tree={
        parent anchor=south,
        child anchor=north,
        fit=band,
      }
      [Causality and Fairness 
        [Pre-Processing
            [Data \\ Manipulation \\ {~\cite{xu2019achieving,salimi2019interventional}}\\
           {~\cite{zhang2017anti,10.5555/3172077.3172438,zhang2018causal}}]
            [Data \\ Integration  \\ {\cite{galhotra2022causal}}]
        ]
        [In-Processing
            [Correcting \\path-specific\\ effects \\ {\cite{kilbertus2017avoiding,nabi2019learning}}\\
            {\cite{chiappa2018causal}}]
            [Fairness under \\ distribution \\shift  \\ {\cite{singh2021fairness, creager2020causal}}]
            [Adversarial\\learning  \\ {\cite{li2021towards}}]
        ]
        [Post-Processing \\ 
            [Constrained \\optimization \\{\cite{kusner2019making,wu2019counterfactual}}  ]
            [Estimators \\{\cite{mishler2021fairness}}]
        ]
      ]
    \end{forest}
    \caption{Segregation of causality-based methods for fairness-aware learning.}
    \label{fig:fairness_tree}
\end{figure}

\noindent Addressing issues of fairness is a prerequisite for applying AI-based learning algorithms to support decisions that critically affect people’s lives, such as offender recidivism, loan approval, disease
diagnosis, hiring, student admissions, etc. A complete understanding of the causal relationships between the sensitive attributes (e.g. gender, race, marital status, etc.)
and the predicted outcome may play 
a crucial role in analyzing and legitimizing the fair or unfair behavior of a learning algorithm.
This section aims to provide an overview of the contributions of causality in existing fairness research, its promises, and how it might be helpful in building fairer models.

\noindent The use of causality to describe and quantify fairness is a distinguishing feature of research conducted in the field of causal fairness. Causality has also proven to be effective in mitigating discrimination. 
We can divide the state-of-the-art causal frameworks for discrimination mitigation into the following categories: (a)  Pre-processing methods, (b) In-processing methods, and (c) Post-processing methods as presented in Figure \ref{fig:fairness_tree}.

\subsection{\textbf{Preliminaries}}
The notions of fairness 
can be categorized into two groups: \textit{individual} and \textit{group}. Group fairness notions assess the large-scale biased effect of the learning algorithm on a certain legally protected group of the underlying dataset. Individual fairness notions measure the difference in the decisions predicted for similar individuals in a population.
Causal fairness notions can be further segregated based on two criteria: 
(a)\textit{ Counterfactual fairness (CF)}\label{CF} and (b) \textit{Interventional Fairness (IF)}\label{IF}.

\subsubsection{Counterfactual fairness (CF) \label{section_cf}} CF measures fairness by quantifying the effect of sensitive attributes on the predicted outcome through counterfactuals. If the sensitive attribute (e.g. $S = \text{`gender'}$) is binary then it could take two values protected, i.e. the disadvantaged group ($p^{+} = \text{`female'}$) and non-protected, i.e. favoured ($p^{-}= \text{`male'}$) group.

\noindent \textbf{Counterfactual fairness}~\cite{kusner2017counterfactual} is achieved by a predictor $Y$ for an individual if the probability of achieving the output $Y = y$ remains the same if the value of sensitive attribute changes from $p^{-}$ to $p^{+}$ as presented in Equation \eqref{eq1}, where $X = V\backslash\left \{ S,Y \right \}$ is the set of all variables except the sensitive and outcome variables.
    \begin{equation}
        \mathbb{P}(y_{p^{+}}|X = x, S = p^{+}) =  \mathbb{P}(y_{p^{-}}|X = x, S = p^{-})
    \label{eq1}
    \end{equation}
    where $P(y_{p^{+}})$ is a short notation for $P(Y=y|do(S =p^{+}))$.
    This individual fairness notion assumes that the effect of sensitive attributes on the decision along all causal paths is unfair. But this may not be true in some cases, e.g. from Figure \ref{fig:fair and unfair}, 
    the direct effect of race on the admission outcome is unfair; however, the indirect effect of race on the outcome through the qualification variable is fair. 
    
\noindent \textbf{Path specific counterfactual fairness}~\cite {chiappa2019path}, in contrast to the counterfactual fairness notion, attempts to remove the causal effects of sensitive attributes on the outcome along only unfair causal paths. For a set of paths $\lambda$, path-specific counterfactual fairness exists if Equation \eqref{eq2} is satisfied, where $\bar{\lambda}$ is the set of remaining paths and $P(y_{p^{+}})$ is a short notation for $P(Y=y|do(S =p^{+}))$.
    
    \begin{equation}
       \mathbb{P}(y_{p^{+}|\lambda, p^{-}|\bar{\lambda}}) = \mathbb{P}(y_{p^{-}}) 
    \label{eq2}
    \end{equation}

\noindent \textbf{PC-fairness}~\cite{wu2019pc} is an additional path-specific counterfactual fairness notion for subgroups not just individuals. Given a set of paths ($\lambda$) and a factual condition $X = x \: (X \epsilon V)$, a predictor $Y$ attains PC-fairness if it satisfies the following criteria:
    \begin{equation}
        \mathbb{P}(y_{p^{+}|\lambda, p^{-}|\bar{\lambda}}|X) = \mathbb{P}(y_{p^{-}}|X),
    \label{eq3}
    \end{equation}
    where $P(y_{p^{+}})$ is a short notation for $P(Y=y|do(S =p^{+}))$.

\noindent \textbf{Counterfactual equalized odds} ~\cite{mishler2021fairness} fairness notion is satisfied by a predictor if the respective counterfactual false positive rates (cFPR) and counterfactual false negative rates (cFNR) of the protected group and non-protected group are equal which is not possible practically. Therefore, we can use approximate counterfactual equalized odds. This notion is satisfied by a predictor if the constraints \eqref{eq35} hold, where $\varepsilon^{+}$ and $\varepsilon^{-}$ are predefined thresholds. $Diff^{+}$ ($Diff^{-}$) is the difference between cFPR (cFNR) of the protected  and the non-protected group as presented in Equation \eqref{eq33}.  
 \begin{equation}
        |Diff^{+}| \leq \varepsilon^{+}, \: \: \: \: \: \: \: \:\varepsilon^{+} \: \: \epsilon \: \: [0,1]
    \:\: ~and~ \:\:
        |Diff^{-}| \leq \varepsilon^{-}, \: \: \: \: \: \: \: \:\varepsilon^{-} \: \: \epsilon \: \: [0,1]
    \label{eq35}
    \end{equation}
    \begin{equation}
        Diff^{+} = cFPR(p^{+}) - cFPR(p^{-})  
    \:\: ~and~ \:\:
        Diff^{-} = cFNR(p^{+}) - cFNR(p^{-})
    \label{eq33}
    \end{equation}

\noindent \textbf{Causal Explanation formula}~\cite{zhang2018fairness} is a causal explanation method that helps in dividing the observed discrimination into three counterfactual effects: direct (DE), indirect (IE), and spurious effects (SE) of sensitive attributes on the outcome. The authors designed a causal explanation formula and decomposed the total variation (TV) into DE, SE, and IE, as presented in the following equation.

    \begin{equation}
        TV_{p^{+}, p^{-}}(Y=y) = |SE_{p^{+}, p^{-}}(Y=y) + IE_{p^{+}, p^{-}}(Y=y|S=p^{-})-DE_{p^{+}, p^{-}}(Y=y|S=p^{-}) |
    \label{eq7}
    \end{equation}
    \if{0}
    \begin{equation}
        TV_{p^{+}, p^{-}}(Y=y) = DE_{p^{+}, p^{-}}(y|S=p^{+}) - SE_{p^{+}, p^{-}}(y)-IE_{p^{+}, p^{-}}(y|S=p^{+})
    \label{eq8}
    \end{equation}
    \fi
    This formula demonstrates that the total discrimination experienced by the individuals with S = $p^{-}$ equals the disparity experienced via SE, plus the advantage lost due to IE, and minus the advantage it would have gained without DE.

\subsubsection{Interventional Fairness (IF)} IF measures fairness by quantifying the effect of sensitive attributes on the predicted outcome by intervening on the protected and non-protected attributes.

\begin{figure}
  \centering
  \subfloat[]{\includegraphics[width=0.35\textwidth]{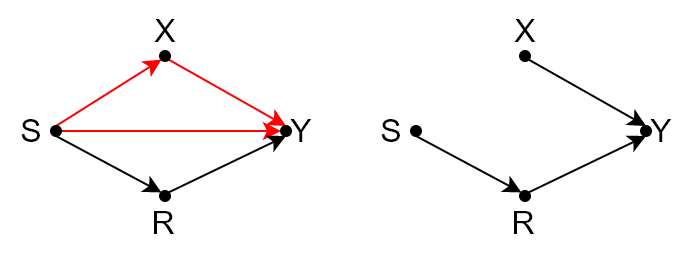}}\hspace{3em}
  \subfloat[]{\includegraphics[width=0.35\textwidth]{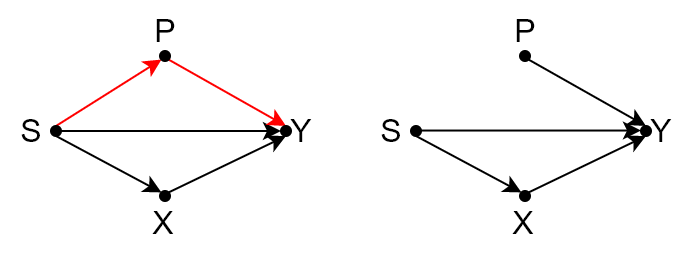}}
  \caption{(a) The left causal graph exhibits unresolved discrimination (along red paths) while the right one is free from unresolved discrimination where $S$ is the sensitive variable, $R$ is the resolving variable, $X$ is the unresolving variable, and $Y$ is the predicted decision. (b) The left causal graph exhibits proxy discrimination (along red paths) while the right one is free from `proxy discrimination' where $S$ is the sensitive variable, $P$ is the proxy variable, and $Y$ is the predicted decision.}
  \label{fig:unresolved and proxy}
\end{figure}

\noindent \textbf{No unresolved discrimination} ~\cite{kilbertus2017avoiding} is a group fairness notion that focuses on the direct and indirect causal influence of sensitive attributes on the decision. It is satisfied when there is no direct path between the sensitive attributes and the outcome, except through a resolving/ admissible variable. In a causal graph, a resolving variable is a variable that is influenced by the sensitive attributes in an unbiased manner. The left causal graph in Figure \ref{fig:unresolved and proxy} (a) exhibits discrimination along the causal paths: $S \rightarrow Y$ and $S\rightarrow X \rightarrow Y$ while the right one is free from discrimination, where $R$ is the resolving variable, $X$ is unresolving variable, and $Y$ is the predicted outcome.

   \if{0}
    \begin{figure}[ht]  
        \begin{minipage}[b]{0.45\linewidth}
        \centering
        \includegraphics[width=\textwidth]{Images/unresolved discrimination.png}
        \caption{Left causal graph exhibiting unresolved discrimination (along red paths) while the right one is free from unresolved discrimination where $S$ is the sensitive variable, $X$ is the resolving variable, $Y$ is the predicted decision.}
        \label{fig:figure1}
        \end{minipage}
        \hspace{0.5cm}
        \begin{minipage}[b]{0.45\linewidth}
        \centering
        \includegraphics[width=\textwidth]{Images/proxy discrimination.png}
        \caption{Left causal graph exhibiting proxy discrimination (along red paths) while the right one is free from unresolved discrimination where $S$ is the sensitive variable, $P$ is the proxy variable, and $Y$ is the predicted decision.}
        \label{fig:figure2}
        \end{minipage}
    \end{figure}
    \fi

\noindent \textbf{Proxy discrimination}~\cite{kilbertus2017avoiding} is another indirect causality-based group fairness notion. It is present in a causal graph when the path between the sensitive attributes and the outcome is intercepted by a proxy variable. A predictor $Y$ avoids proxy discrimination if, for a proxy variable $P$, Equation \eqref{proxy_disc} holds for all potential values of P ($p_{1}, p_{2}$). A proxy variable has the same influence on the outcome as the influence of the sensitive attributes. The left causal graph in Figure \ref{fig:unresolved and proxy} (b) exhibits discrimination along the causal paths: S$\rightarrow$P$\rightarrow$Y; where $P$ is the proxy variable and $Y$ is the predicted outcome. 
        \begin{equation}
        \mathbb{P}(Y=y|do(P = p_{1})) = \mathbb{P}(Y=y|do(P = p_{2}))\:  \forall \: p_{1},p_{2} \: \epsilon\: dom(P).
        \label{proxy_disc}
    \end{equation}

\noindent \textbf{Total Effect} ~\cite{pearl2009models} is the causal version of the statistical parity group fairness notion. It measures the effect of changing sensitive attribute values on the outcome along all causal paths from sensitive attributes to the outcome as presented in Equation \eqref{eq9} 

    \begin{equation}
        TE_{p^{+}, p^{-}}(Y=y) = \mathbb{P}(y_{p^{+}}) - \mathbb{P}(y_{p^{-}}) 
    \label{eq9}
    \end{equation}

\noindent \textbf{Individual direct discrimination} ~\cite{zhang2016situation} identifies direct discrimination at individual level. In any classification task, an individual is compared with $n$ similar individuals sought from the protected  group, denoted as $D$ and $n$ similar individuals from the non-protected group, denoted as $\mathbf{\bar{D}}$.  The similarity between the two individuals $(i, i')$ is measured using causal inference as presented in Equations \eqref{eq11} and \eqref{eq12}, where $CE$ is the causal effect of each selected variable $(x_{k})$ on outcome variable, $VD$ is the distance function, and range is the difference between the maximum and minimum of the variable $x_{k}$. 
    The individual is deemed not to be discriminated against if the difference between the positive prediction rates for the two groups $(\textbf{D}, \mathbf{\bar{D}})$ is under a predefined threshold. 
    \begin{equation}
        d(i,i') = \sum_{k=1}^{|X|}|CE(x_{k},x_{k'}) \cdot VD(x_{k},x_{k'})| 
    \label{eq11}
    \end{equation}
    \begin{equation}
        CE(Y=y)=\mathbb{P}(Y=y|do(X))-\mathbb{P}(Y=y|do(x_{k}^{'},X\backslash x_{k}))
     \:~~~~and~~~~\:   VD(x_{k},x_{k'}) = \frac{|x_{k}-x_{k'}|}{range}
    \label{eq12}
    \end{equation}
    
\noindent \textbf{Equality of effort}~\cite{huan2020fairness} assesses discrimination by measuring the amount of effort required by the marginalized individual or group to reach a certain level of the outcome as shown in Equation \eqref{eq155}.  The minimal effort required to achieve the $\gamma$-level of outcome is computed using Equation \eqref{eq14},
 where $G_{+}$ and $G_{-}$ represent the set of individuals with S = $p^{+}$ and S = $p^{-}$ respectively which are similar to the target individual, ${E}[Y_{G^{+}}^{t}]$ is the expected value of outcome under treatment T = t for the set $G_{+}$. This notion of fairness is based on potential outcomes framework. 
    \begin{equation}
        \psi_{G^{+}}(\gamma) = \psi_{G^{-}}(\gamma) \\
    \label{eq155}
    \end{equation}
    \begin{equation}
        \psi_{G^{+}}(\gamma) = argmin_{t\epsilon T}\mathbb{E}[Y_{G^{+}}^{t}] \geq \gamma  
    \label{eq14}
    \end{equation}

\noindent \textbf{Interventional and justifiable fairness} ~\cite{salimi2019interventional} are stronger versions  
    of the Total Effect fairness notion.  
    The total effect intervenes on the sensitive attribute; however, interventional fairness intervenes on all attributes except the sensitive attribute.  A classification algorithm is interventionally K-fair if for any assignment of K = k and output Y = y the following equation holds. K is a subset of attributes (V) except the sensitive attribute (S) and the outcome variable ($K \subseteq V\setminus \left\{S, Y \right\}$ )
    \begin{equation}
        \mathbb{P}(y_{p^{+},k}) = \mathbb{P}(y_{p^{-},k})
    \label{eq15}
    \end{equation}
    Justifiable fairness is a special case of interventional fairness, where we only consider those attributes for intervening that are admissible/resolving (E) or a superset of admissible variables:
    \begin{equation}
        \mathbb{P}(y_{p^{+},k}) = \mathbb{P}(y_{p^{-},k}), k\supseteq E. 
    \label{eq16}
    \end{equation}

\noindent \textbf{Causal fairness} ~\cite{galhotra2022causal} identifies a classifier as fair if for any given set of admissible variables E, the following equation holds:
    \begin{equation}
        \mathbb{P}(y_{p^{+},e}) = \mathbb{P}(y_{p^{-},e}), e\subseteq E. 
    \label{eq17}
    \end{equation}

\subsection{\textbf{Pre-processing Methods}} 
Pre-processing methods are considered to be the most generalizable methods. These methods intend to manipulate the dataset in order to make it bias-free before feeding it to any learning algorithm. 

\noindent \textbf{Data Augmentation:}~\citet{10.5555/3172077.3172438} calculate path-specific effects of sensitive attributes on the predicted outcome and compare them to a predefined threshold $\tau$. If the calculated path-specific effect exceeds $\tau$, this indicates the presence of direct and indirect discrimination. Later they eliminate both direct and indirect discrimination by generating a bias-free dataset through causal network manipulation that guarantees path-specific effects under $\tau$.~\citet{zhang2018causal} further 
identify and handle the situations in which indirect discrimination cannot be measured because of the non-identifiability of certain path-specific effects. In such cases, the authors suggest setting an upper and lower bound on the effect of indirect discrimination. Another discrimination discovery and prevention causal framework is proposed by~\citet{zhang2017anti}. In this work, the authors detect direct and indirect system-level discrimination by measuring the path-specific causal effects of sensitive attributes on the outcome. To prevent discrimination, they modified the causal network to generate a new bias-free dataset.
\citet{xu2019achieving} has proposed a utility-preserving and fairness-aware causal generative adversarial network (CFGAN)\label{CFGAN} to generate high-quality and bias-free data.  
~\citet{salimi2019interventional} detected discrimination using interventional and justifiable fairness notions. To eliminate discrimination, they used causal dependencies between sensitive attributes and outcome variables to add and remove samples from the training data.

\noindent \textbf{Data Integration:} Data integration aims to combine data from various sources that capture a comprehensive context and enhance predictive ability.~\citet{galhotra2022causal} modeled the problem of ensuring {\em causal fairness} in a learning task as a fair data integration problem by combining additional features with the original dataset. They proposed a conditional testing-based feature selection method that guarantees high predictive performance without adding bias to the dataset.
\subsection{\textbf{In-processing Methods}}
In-processing methods achieve fairness by altering the learning algorithm.    

\noindent \textbf{Correcting path-specific effects:} The sensitive attributes can affect the outcome through both fair and unfair causal pathways as explained in Section \ref{causal_graph}.  
~\citet{kilbertus2017avoiding}  proposed to deal with such a situation by constraining the parameters of the learning algorithm so that the causal effects along both fair and unfair causal paths from sensitive attribute to outcome variable is removed. They used ``proxy discrimination'' and ``unresolved discrimination'' fairness notions to detect discrimination.  \cite{nabi2019learning} proposed to deal with such situations by constraining the path-specific effect during model training within a certain range. The methods proposed by~\cite{kilbertus2017avoiding,nabi2019learning} cater for the causal effects of sensitive attributes on the outcome without distinguishing between fair and unfair causal effects, thus negatively impacting the predictive performance of the learning algorithm. A solution to this problem is proposed by~\citet{chiappa2019path}. The author presented a causal framework that ensures path-specific counterfactual fairness by correcting the observations of such variables that are descendants of sensitive attributes along only unfair causal paths so that only individual unfair information is eliminated while the individual fair information is retained, hence improving predictive performance of the framework. 

\noindent \textbf{Fairness under distribution shift:} The problem of learning fair prediction models with covariates distributed differently 
in the test set than in the training set is studied by~\citet{singh2021fairness}. They proposed a method based on feature selection to achieve fairness given the ground truth graph that explains the data. Fairness concerns also surface when AI-based learners deal with dynamically fluctuating environments and produce long-term effects for both individual and protected groups.~\citet{creager2020causal} have proposed that in such dynamical fairness setting when the dynamic parameters are unknown, causal inference can be utilized to estimate the dynamic parameters and improve off policy estimation from historical data. 
 
\noindent \textbf{Adversarial learning:}~\citet{li2021towards} proposed an adversarial learning-based approach to achieve the goal of personalized counterfactual fairness for users in recommendation systems. They attempt to remove sensitive features information from the user embeddings by using a filter module and a discriminator module to make the learner's decisions independent of the sensitive features. 
\subsection{Post-processing Methods} These methods tailor the outputs of the learner to achieve fair outcomes. 

\noindent \textbf{Constrained optimization:}
~\citet{wu2019counterfactual} proposed a method to bound the unidentifiability of counterfactual quantities and used c-component factorization to identify its source.  They proposed a graphical criterion to determine the lower and upper bound on counterfactual fairness in unidentifiable scenarios. Finally, they proposed a post-processing method to reconstruct the decision model to achieve counterfactual fairness.
Similarly,~\citet{kusner2019making} achieved counterfactual fairness by constraining the beneficial effects obtained by an individual under a limit depending on the sensitive attribute of the individual.

\noindent \textbf{Doubly robust estimators:}~\citet{mishler2021fairness} proposed a post-processed predictor, estimated using doubly robust estimators, to achieve the counterfactual equalized odds fairness notion. Through experiments, they also proved that their method has favorable convergence properties. 

\subsection{Conclusion}

A unique trait of  fairness research is the usage of multiple metrics to define/measure it. Consequently, choosing the most appropriate notion of fairness applicable to a particular situation is an important task.
On the other hand, even if a fairness notion is found suitable for a scenario, it may not be applicable due to the problem of identifiability, as Pearl's SCM framework requires causal quantities, counterfactuals, and interventions, to be identifiable ~\cite{makhlouf2020survey}.
Most of the discrimination mitigation approaches discussed above  to achieve the goal of causal fairness are based on the synthesis of a bias-free dataset. These methodologies are only applicable in a static environment where all data are available in advance. An interesting future direction could be the extension of such methods for online learning, where not all data are available beforehand~\cite{galhotra2022causal}.  Most of the causality-based fairness solutions discussed above rely on the assumption that the underlying data is independent and identically distributed (IID)\label{IID}. However, real-world use cases include non-IID data, therefore, another future direction could be to design causality-based decision support systems which relax the assumption of non-IID data and provide non-discriminatory predictions. Finally, all research done in the field of causal fairness is connected to classification tasks~\cite{zhang2017anti}, it will be interesting to expand it to 
achieve causal fairness in community detection, word embedding, named entity recognition, representation learning, semantic role labeling, language models, and machine translation~\cite{li2021towards}.

\section{Causality and Robustness}
\label{sec:robustness}

\begin{figure} 
  \centering   
  \small
    \begin{forest}
      for tree={
        parent anchor=south,
        child anchor=north,
        fit=band,
      }
      [Causality and Robustness
        [Pre-Processing\\
            [Data\\ 
             Augmentation\\
               ~\cite{kaushik2020explaining, Ilse_2021_simulatinginterventions, Mao_2021_generative, little2019causal}
            ]
            [Problem\\
            Abstraction\\
               ~\cite{Wang_2022_CDL, ilyas2022datamodels}
            ]
        ]
        [In-Processing\\
          [Objective\\
          (Loss) Design\\
           ~\cite{arjovsky2019invariant, dominguez2022adversarial,lu2021invariant, Mitrovic_2020_relic}\\
           ~\cite{zhang2021causaladv, teney2020learning, wang2021enhancing}
          ]
          [Architecture\\
          Design\\
           ~\cite{goyal2021recurrent, liu2022towards, zhang2020causal}
          ]    
        ]
        [Post-Processing\\
            [Altering\\
             Predictions\\
           ~\cite{Chen_2021_posthoc, wang2022ISR}
            ]
            [Model\\
             Selection\\
            ~\cite{kyono2019improving}
            ]
        ]
      ]
    \end{forest}
    \caption{Structure of approaches introducing causality in robustness.}
    \label{fig:robustness_tree}
\end{figure}
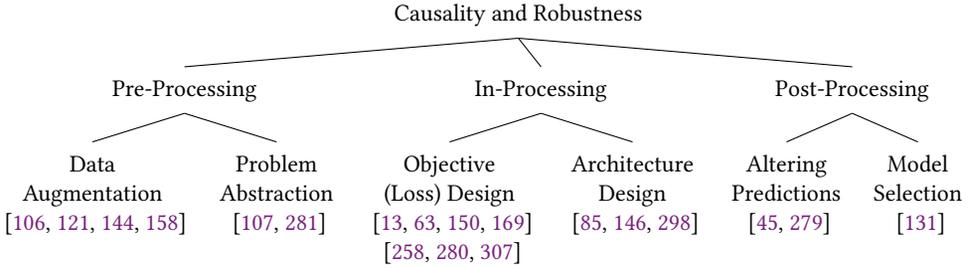

Training of modern machine learning (ML) systems builds upon the assumption that all observations - whether training or test data - are \textit{independent and identically distributed} (IID) under a single distribution.
As it is improbable that training data can perfectly represent the sample distribution of real-world data, striving for more robust ML systems is of utmost importance.
Robust behavior is not only required for safety-critical applications of ML but also essential for the trustworthiness of AI. 
We believe that AI systems that are error-prone to small changes in the working environment will not be able to gain the complete trust of humans.
Even high-performing ML models often cannot differentiate between \textit{style variables}, which are content-independent and irrelevant information for the task, and the information-rich, relevant, and invariant \textit{content variables}~\cite{Schoelkopf_2022, kaddour_2022_causalmlsurvey}.
Causal information about the problem can provide models with a better understanding of the task and eliminate such spurious correlations.
Inspired by this perspective, many researchers in recent years investigated the connection between robust machine learning and causality. 
We summarize related publications and organize them into three broad categories: (i) \textbf{pre-processing}, (ii) \textbf{in-processing}, and  (iii) \textbf{post-processing} methods. 
Figure \ref{fig:robustness_tree} provides an overview of the techniques discussed in this section.

\subsection{Preliminaries} We briefly introduce the most important concepts for this section, including a high-level definition of robustness and the primary sources of non-robust behavior.

\noindent{\textbf{Robustness:}} The notion of robustness boils down to the question of how sensitive the ML model's output is to changes in the input. Minor changes in the input should not significantly alter the performance of robust AI systems. Instead, performance should degrade \textit{gracefully}: slowly and gradually with the deviation in input distribution.
We distinguish between naturally occurring and artificially crafted distributional shifts. The first type of shift is represented by the notion of \textit{out-of-distributional} data, whereas the second type of shift is represented by \textit{adversarial examples}.

\noindent{\textbf{Out-Of-Distribution Data:}}
Out-Of-Distribution (OOD) data represents naturally occurring data with previously unseen characteristics. For instance, computer vision models trained to solve MNIST classification may encounter naturally perturbed images (e.g. numbers written in a different orientation or a different color). Such perturbations represent data points from a different, shifted distribution and, therefore, may lead to poor performance of our computer vision model. These data points are, thus, considered to be \textbf{out-of-distribution}. Methods such as \textit{Data Augmentation} (e.g. methods discussed in~\cite{Shorten_2019_Data_Augmentation}) can increase robustness towards OOD-data by providing models with additional, naturally perturbed data via hand-crafted rules~\cite{Schoelkopf_2021, teney2020learning}. These methods, however, cannot cover all possible environmental settings~\cite{teney2020learning}.

\noindent\textbf{Adversarial Examples:}
Another hurdle in robust machine learning is the continuous rise of \textit{Adversarial Examples} (AE). AEs represent artificially perturbed input values intending to fool machine learning models. Such examples are especially concerning for the field of Trustworthy AI, as these perturbed input values look benign to humans. Despite the existence of various developed defenses against continuously evolving AEs, research into improved attacks and defenses continues to this day. We refer to \cite{Akhtar_2021_Survey} for a recent overview of such methods in the computer vision domain. 

\subsection{\bf Pre-Processing}
\label{subsub:robustpreproc}
Pre-processing methods built upon causality are mostly \textit{Data Augmentation} methods, which create causally motivated augmentations. 
We will also discuss alternative, exciting approaches to pre-processing that we cover under the umbrella term \textit{Problem Abstraction}.

\subsubsection{\bf Data Augmentation}
Data Augmentation is the most common pre-processing method to induce causality.
There are several methods~\cite{Mao_2021_generative, Ilse_2021_simulatinginterventions}
which use the notion of causal graphs to motivate data augmentation. 
For instance,~\citet{Ilse_2021_simulatinginterventions} try to find and apply transformations that emulate the intervention on high-level domain-specific features (e.g., the orientation of handwritten digits) within data points. Such information is only spuriously correlated to the output label and, as such, should not affect the decision-making of ML models. To find such a transformation without an SCM, the authors propose to train a classifier that can predict the domain of data points (e.g., the given number is rotated by 60°). They then choose the augmentation of a pre-defined set of transformations that leads to the \textit{lowest} accuracy of the domain classifier. Applying the selected augmentation "destroys" the most domain-specific information. Such augmented training data, in turn, reduces the likelihood of ML models overly relying on domain-specific features that are only spuriously correlated to the label. 
 \citet{Mao_2021_generative} show that it is also possible to generate intervention-simulating data via GANs by identifying \textit{interpretable controls} through GANSpace. 
 Manipulating the data generated through these controls is equivalent to intervening in the underlying SCM. Alternatively, one can generate counterfactual examples by augmenting the data just enough to flip the label. Following this principle,~\citet{kaushik2020explaining} developed a human-in-the-loop process that improves robustness on NLP tasks compared to alternative, non-causal methods.
As shown by~\citet{little2019causal}, it is also possible to induce causality through bootstrapping. The authors developed \textit{causal bootstrapping}, which utilizes information provided by a causal graph to sample data whose deducible observations better reflect the domain's causal relationships. Models trained on causally sampled data demonstrate increased robustness against spurious correlations.

\subsubsection{\bf Problem Abstraction}
Instead of data sampling, some methods try to abstract and simplify the problem. One example is the \textit{Datamodeling}~\cite{ilyas2022datamodels} framework, which allows researchers to approximate the behavior of large and complex models on the given data through a set of simple linear functions. 
\citet{Wang_2022_CDL} simplify the problem for reinforcement learning (RL) agents using information encoded in a causal graph. 
The authors propose to create state abstractions for RL agents that only contain the relevant one-to-one causal dependencies between variables and actions.
RL agents that utilize this state abstraction exhibit higher robustness towards unseen states, cover a wider range of tasks than agents trained with other methods and demonstrate higher sample efficiency.

\subsection{\bf In-Processing}
\label{subsub:robustinprocess}
 
It is possible to induce notions of causality as part of the algorithm either 
through the definition of a \textit{causality-aware optimization objective} or via \textit{architectural design choices}.

\subsubsection{\bf Objective (Loss) Design}
Most causal in-processing techniques incorporate an optimization objective (e.g. a loss function or a regularization term) that guides ML models to a more causality-aware behavior. We will introduce three possible causal foundations for such objectives.

\noindent\textbf{Content and Style Variables:}~\citet{zhang2021causaladv} use a causal graph to model the generation process of AEs to examine the source of adversarial vulnerability. They conclude that AEs 
exploit the spurious correlations between style variables and labels to mislead classifiers. However, the \textit{adversarial distribution} is drastically different from the natural one. Consequently, the authors developed a loss function that \textit{aligns} the two distributions. Compared to other robust classifiers, aligned classifiers demonstrate higher accuracy on adversarial data without significantly worse performance on natural data.
\citet{wang2021enhancing} developed a regularization term for \textit{logistic regression} models allowing researchers to penalize causal and spurious features separately. This form of regularization explicitly requires researchers to categorize features as either causal, spurious, or remaining (not identified as either of the two). Given such information, models optimized with the term showcase improved robustness on lower and higher-dimensional data.

\noindent\textbf{Multiple Environments:} One vital insight for causal ML is the connection between the causal relevance of a feature and its invariance across environments~\cite{peters2016causal}. 
The basic idea is that style variables (e.g. the image background) greatly vary across environments (i.e. unique experimental settings), whereas content variables (e.g. an animal's anatomy) remain \textit{invariant}. Consequently, guiding models to perform well across environments should lead to models capable of differentiating between content and style variables, which, in turn, makes them more robust.
The most prominent example is the loss function \textit{invariant risk minimization} (IRM)~\cite{arjovsky2019invariant} that empirically leads to higher robustness.
\citet{Mitrovic_2020_relic} were able to transfer this premise to the Self-Supervised Learning (SSL) setting to improve OOD performance. The resulting SSL objective requires generated data representations to be stable across different interventions simulated by data augmentation.
On the assumption that ``the prior over the data representation belongs to a general exponential family when conditioning on the target and the environment''~\cite{lu2021invariant}, the \textit{iCaRL} framework is able to outperform IRM, but causal discovery is needed to identify the causally-relevant latent variables.

\noindent\textbf{Counterfactuals:} Researchers also successfully developed counterfactual-based loss functions.~\citet{teney2020learning} designed an auxiliary loss for supervised learning that gives additional attention to pairs of data that are counterfactuals of one another.
Raising awareness for counterfactuals can also increase the robustness of already causal methods, such as recommendation via Causal Algorithmic Recourse. Such algorithmic recourse systems try to find minimal-costly actions that result in a counterfactual representing a desirable outcome.~\citet{dominguez2022adversarial} were able to further enhance this framework by also accounting for uncertainties stemming from adversarially perturbed features. However, instead of considering all possible perturbations within an $\epsilon$-range, the authors solely considered perturbations within the (SCM-guided) \textit{counterfactual neighborhood} (i.e., only instances in $\epsilon$-range that represent counterfactuals to the given data).  

\subsubsection{\bf Architecture Design}
\label{par:robustdesign} 
\citet{goyal2021recurrent} try to take advantage of the \textit{independent causal mechanisms} (ICM) principle. It states that the causal dynamics of a domain are built upon ``autonomous modules that do not inform or influence each other.''~\cite{Schoelkopf_2022}.
They achieve this by implementing a sequential architecture of independently acting recurrent subsystems that only communicate sparsely with one another. Each subsystem is designed to emulate a mechanism of the causal generative process in the hope of better capturing the domain's causal structure. The resulting architecture is more robust to distributional shifts than, e.g. LSTM or Transformers.

\noindent An alternative, more feature-focused architecture is \textit{deep Causal Manipulation Augmented Model} (deep CAMA)~\cite{zhang2020causal} - a deep generative model whose design is consistent with the causality encoded in a given causal graph. It not only considers the effect of the output label on the input data but also considers manipulable variables (e.g., rotation and color of MNIST digits) and non-manipulable variables (e.g., handwriting style of MNIST digits) influence on the input. The authors achieve this by adding autoencoders for both the variables. This design choice allows the resulting generative model to better distinguish between relevant/causal features and non-relevant ones. Deep CAMA showcased improved robustness against adversarial attacks on MNIST data.

\noindent \citet{liu2022towards} follow a similar approach to designing a robust motion forecasting model. The authors argue that latent variables of the motion forecasting tasks are either (i) \textit{invariant variables} such as laws of physics, which are crucial for correct motion forecasting, (ii) \textit{hidden confounders} like the motion style that can sparsely differ between environments but still need to be considered for optimal performance, or (iii) \textit{non-causal spurious features} that can drastically vary between environments. Based on this categorization, the authors developed an architecture that provides higher robustness toward style shifts.

\subsection{\bf Post-Processing}
\label{subsub:robustposthoc}

The causality framework also enables researchers to impact the robustness of their ML-pipeline \textit{after the training phase}. Post-processing methods discussed in this section either directly alter the predictions of a given trained
model or enable a causality-informed selection between a set of models. 

\subsubsection{\bf Altering Predictions}
\textit{Counterfactual regularization}~\cite{kaddour_2022_causalmlsurvey}, which tries to remove the confounding effects of unobserved variables, is a common approach to instilling causality through post-processing.
One can achieve this by first estimating the effects of unobserved variables, referred to as the \textit{counterfactual prediction}~\cite{Chen_2021_posthoc}.
By then taking the difference between the counterfactual prediction and the \textit{factual prediction} (built upon both causally relevant and spurious information), a deconfounded prediction can be extracted. Using such a counterfactual regularization,~\citet{Chen_2021_posthoc} successfully improved the quality of trajectory predictions in multi-domain settings.
\citet{wang2022ISR} introduce an alternative approach called \textit{Invariant-Feature Subspace Recovery} (ISR). The authors first extract the feature representation of the given data via the model's hidden layers. They then recover the subspace spanned by invariant features (i.e., content variables), fit a linear predictor in the resulting manifold, and substitute the model's classification layer with the subspace predictor. Post-processing with ISR improved the performance of trained models on multiple OOD-datasets.

\subsubsection{\bf Model Selection}~\citet{kyono2019improving} introduce a pipeline that utilizes (incomplete) causal information in the form of a DAG in a post-hoc manner. Given a set of trained models and the data these models used, it is possible to score models based on how well the model's predictions abide by the ``rules'' induced by the given causal structure (DAG). The authors propose to choose the model whose predictions follow the given causal relationships the most. Classifiers chosen via this causal model selection technique empirically showcase higher robustness in OOD-learning settings.

\subsection{\bf Conclusion} 
Despite the relative novelty of causal learning, researchers have already applied the notions of causality to domains such as computer vision, NLP, and recommendation. Naturally occurring distributional shifts are the focus of these advancements, though causal learning can also increase robustness towards AEs. The success of causality in fields outside of traditional supervised learning, e.g. reinforcement 
learning~\cite{Wang_2022_CDL} and self-supervised learning~\cite{Mitrovic_2020_relic} further show the legitimacy of improving robustness via causal learning.
We see multiple exciting avenues for future studies into robust causal ML. 
For instance, researchers need to be aware that OOD-datasets may vastly differ in the type of distributional shifts they emulate~\cite{Ye2021ood}. Hence, causal solutions for robust ML need to be analyzed and compared to other (both causal and non-causal) state-of-the-art approaches on benchmarks that contain a diverse collection of datasets.
It will also be interesting to (further) explore related fields such as \textit{Neurosymbolic AI} (where researchers enhance ML systems through the use of knowledge-based systems) or \textit{Object-Centric Learning} (a special case of Causal Representation Learning~\cite{Schoelkopf_2021} where visual scenes are modeled as compositions of objects that interact with one another). We also believe that research into causal solutions for \textit{certified robustness} and \textit{concept drifts} occurring in online learning will lead to interesting new solutions.

\section{Causality and Privacy}
\label{sec:privacy}

In Section~\ref{sec:robustness}, we discussed the robustness aspect of AI from the viewpoint of the model and noted out-of-domain generalization as a fundamental challenge. The problem becomes even more challenging in a distributed learning scenario, where the AI model gathers data from different users for training. In such a setting, it must be ensured that the users' private data is not exposed. Recent studies~\cite{yeom2018privacy, tople2020alleviating} have demonstrated that weak generalizability of a model could be exploited to design attacks that can expose users' private information (e.g., whether the user's data were used in training the model). With their inherent ability to generalize to out-of-distribution (OOD) data, causal models can help in preventing such attacks.     

\subsection{\bf Preliminaries}

\noindent{\bf Learning paradigm}: Machine learning algorithms are often trained on a large amount of training data, which can contain sensitive information. We consider a particular setting where data are collected from one or multiple users to train a machine-learning model. For example, the data could be collected from users' mobile devices to train a better speech recognition model. 
The most relevant variant in the context of privacy is \textbf{federated learning}~(FL)~\cite{mcmahan2017communication}. FL proposes to learn a global  model without collecting users' data into a central server, rather keeping the data in users' devices. Data are processed locally to update a local model, while intermediate model updates are sent to the central server, which are then aggregated to update the global model. Once updated, the global model is sent to the local devices.
In this part, we will specifically look into attacks on federated learning as it involves dealing with users' data where privacy is of utmost importance.

\noindent{\bf Privacy attacks}: 
There has been a growing body of work aimed at designing different types of attacks against FL systems (refer to~\citet{jere2020taxonomy} for a comprehensive overview).
However, one of the most common types of attack posed to FL systems and where causality-driven models have been deployed more often is \textbf{membership inference attack}~\cite{shokri2017membership}. In this type of attack, the attacker tries to determine whether a particular data point was used in training the model, with only model predictions available to the attacker~\cite{nasr2018comprehensive, yeom2018privacy}. The core idea is to exploit the stochastic gradient descent (SGD) algorithm to extract information about a client's data. 
The attacks can be initiated both from the server's side and the client's side. For a comprehensive review of membership inference attacks on FL, refer to \citet{nasr2018comprehensive}. 
Membership inference attack has been linked to model generalizability~\cite{nasr2018comprehensive}, and causal models are adept at improving the generalizability of models. Hence, the bulk of work deploying causal models against privacy attacks has concentrated on combating membership inference attacks, which we discuss in detail in this section. \emph{Differential privacy} (DP)~\cite{dwork2006our} 
ensures that the presence or absence of a data point does not significantly influence the output. Hence, a defense mechanism against membership attacks should provide better differential privacy guarantees.

\noindent{\bf Evaluation}: The success of a membership inference attack is measured in terms of accuracy on the binary classification task of determining membership~\cite{shokri2017membership}. Similarly, \citet{yeom2018privacy} introduce \emph{advantage}, which is measured as the difference in true and false positive rates in membership prediction. Hence, the goal of any defense mechanism against membership attacks is to reduce the attack's accuracy or advantage. 


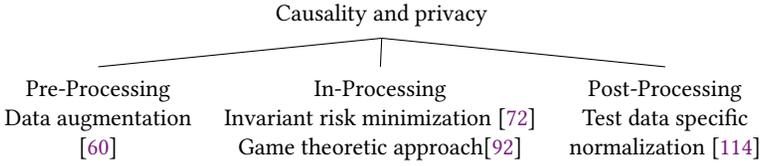
\begin{figure} 
  \centering  
  \small
    \begin{forest}
      for tree={
        parent anchor=south,
        child anchor=north,
        fit=band,
      }
      [Causality and privacy
          [Pre-Processing \\ Data augmentation \\ \cite{de2022mitigating}]
          [In-Processing  \\ Invariant risk minimization \cite{francis2021towards} \\
          Game theoretic approach\cite{gupta2022fl}]
          [Post-Processing \\ Test data specific \\ normalization \cite{jiang2021tsmobn}]
      ]
    \end{forest}
    \caption{Causality and privacy approaches.}
    \label{fig:privacy_tree}
\end{figure}

\noindent{\bf Causal models for improving privacy}:
\citet{yeom2018privacy} demonstrate that model overfitting significantly contributes to the  leakage of membership information.  
For example, consider a model $A$, a data point $z$, and a loss function $l \leq B$ where $B$ is a constant. An attack strategy is to first query the model (i.e., computing $A(z)$) and then output that $z$ as a member of the training set with probability $1 - l(A, z)/B$. For such an attack, the performance in determining membership is shown to be proportional to the generalization error (i.e., the extent of overfitting).
Hence, the defense mechanisms propose to deploy model generalization techniques such as learning rate decay, dropout~\cite{salem2018ml} or adversarial regularization~\cite{nasr2018machine}. However, these mechanisms assume that the train and the test datasets are sampled from the same distribution, which is not always the case, particularly for FL. In fact, \citet{tople2020alleviating} argue that vulnerability to such attacks is exacerbated when a model is deployed on unseen data. They further utilize the generalization property of causal models and establish a theoretical link between causality and privacy. It is further argued that the models learned through causal features generalize better to distribution shifts and provide better privacy guarantees than equivalent association models. 
This has resulted in a growing body of work exploring causal models against privacy attacks~\cite{tople2020alleviating,  francis2021towards,jiang2021tsmobn,chandrasekaran2021causally,gupta2022fl,de2022mitigating}.
We segregate these
methods into \textbf{pre-}, \textbf{in-} and \textbf{post-}processing methods (refer to figure~\ref{fig:privacy_tree}). 

\subsection{\textbf{Pre-processing }} 
The key idea here is to manipulate the training data at each client, which would result in better generalization.
\citet{de2022mitigating} introduce a causality-based data augmentation to mitigate the problem of domain generalization in FL. The authors argue that data augmentation can reduce the heterogeneity across user data distribution, thereby making them more similar for the server model. The proposed data augmentation method is based on the notion of structural causal model~(SCM) as described in section~\ref{sec:scm}. The authors posit that a data point $X_i$ is generated by a common cause $Z$ and some random variation $\epsilon_i$. Following SCM, this could be represented as $X_i \coloneqq g(Z, \epsilon_i)$ where $g$ is the causal mechanism that remains invariant. An augmented data point $\hat X_i$ could then be generated by defining a transformation $\tau$ over $\epsilon_i$ and then following the SCM $\hat X_i \coloneqq g(Z, \tau(\epsilon_i))$.

\subsection{\textbf{In-processing}} This class of methods aims at learning domain invariant features while training.
\citet{francis2021towards} propose to collaboratively learn causal features common to all collaborating users/clients through invariant risk minimization~\cite{arjovsky2019invariant} (introduced in Section \ref{subsub:robustinprocess}). 
The proposed method is able to defend better against membership inference and  property inference attacks  in comparison to the vanilla FL algorithms. \citet{gupta2022fl} propose FL games, a game theoretic method, to solve the problem of generalization in FL. The key idea is to learn causal representations that are invariant across users. Each client serves as a player that competes to optimize its local objective, while the server guides the optimization to a global objective. An equilibrium is reached when all the local models across users become equivalent, thereby achieving generalization across users' data distributions as well as finding invariant representations. Although the authors do not explicitly demonstrate the effectiveness of their model against specific attacks, it can be argued that better generalization should lead to improved robustness against inference attacks. 
Given these methods aim to discover a set of invariant features that directly influence the outcome (hence speculated as causal), they can be loosely placed under causal discovery. However, it is difficult to conclude whether the discovered features are indeed causal.

\subsection{\textbf{Post-processing }} 
\citet{jiang2021tsmobn} argue that the FL training could be represented as a structural causal model with four variables, input data ($X$), raw extracted features ($R$), normalized features ($F$) (obtained by applying batch normalization~\cite{ioffe2015batch}) and the output ($Y$), following a causal structure $X \rightarrow R \rightarrow F \rightarrow Y$.
Although heterogeneity of data can lead to each client fitting its individual feature distribution as opposed to the global objective, batch normalization (BN) layers normalize the training data to a uniform distribution, thereby allowing the global model to converge. 
However, when dealing with unseen test data ($D^u$), BN layers with the estimated training statistics run the risk of normalizing it improperly, thereby introducing an edge $D^u \rightarrow F$ and making $D^u$ a confounder. Performing a causal intervention by introducing a surrogate variable $S$, which is test-specific statistics of raw features $R$ (i.e., BN normalization parameters are now computed from the test set instead of the training set),  
leads to blocking the path between $D^u$ and $F$ and getting rid of the confounder.  
This achieves better generalization and hence better robustness against attacks.

\subsection{Conclusion}
In this section, we presented an overview of various attempts at improving defenses against privacy attacks on FL systems through causality-driven methods.
Most of them have  exploited the OOD generalization capabilities of causal methods 
as this successfully defends   against inference attacks, which calls for similar investigations with respect to other types of attacks.
Except for \citet{tschantz2020sok}, who propose to apply results from causality theory while studying differential privacy (DP)~\cite{dwork2006our}, this has largely remained unexplored and might be another interesting line of future research.

\section{Causality and Auditing (Safety and Accountability)}\label{sec:safety}

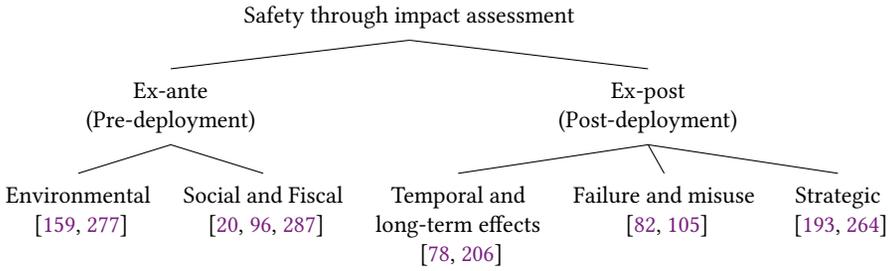
\begin{figure} 
  \centering 
  \small
    \begin{forest}
      for tree={
        parent anchor=south,
        child anchor=north,
        fit=band,
      }
      [Safety through impact assessment
        [{\hyperref[]{Ex-ante}}\\ (Pre-deployment)\\
            [{\hyperref[]{Environmental}}\\
               {~\cite{mareddy2017environmental,voegeli2019sustainability}}
            ]
            [{\hyperref[]{Social and Fiscal}}\\
               {~\cite{becker2001social,wu2015causality,haseeb2019economic}}
            ]
        ]
        [{\hyperref[]{Ex-post}}\\ (Post-deployment)\\
            [{\hyperref[]{Temporal and}}\\long-term effects\\
               {~\cite{ramaciotti2021auditing,garimella2017long}}
            ]
            [{\hyperref[]{Failure and misuse}}\\
               {~\cite{ibrahim2021causality,gillespie2021impact}}
            ]
            [{\hyperref[]{Strategic}}\\
               {~\cite{patro2022fair,tsirtsis2020decisions}}
            ]
        ]
      ]
    \end{forest}
    \caption{Safety through (causal) impact assessment.}
    \label{fig:safety_tree}
\end{figure}

Automated systems often enabled by AI, 
when deployed at scale, interact with people, each other, and also with other ecosystem or environmental parameters, thereby having a potential for widespread unforeseen (desirable or undesirable) effects.
Some undesirable effects can cause significant societal harm, such as the creation of online filter bubbles and polarization due to personalized consumption-centric information retrieval systems powered by various ML models~\cite{pariser2011filter}.
Following such effects of AI, there has been a lot of interest in finding potential negative effects before deployment or on the run and make necessary changes to avoid or neutralize such effects.
\citet{selbst2021institutional} describes Algorithmic Impact Assessment (AIA) as a regulatory strategy for addressing and correcting algorithmic harms.
EU regulatory guidelines also emphasize the need for environmental, social, and economic impact assessments before deployment.
The phrase `impact assessment' itself suggests a causal relationship between system design and impact.
Thus, causal inference has long been used for impact assessment of algorithmic and non-algorithmic systems. Depending on when an assessment occurs, it can be categorized into (i) ex-ante (before deployment) and (ii) ex-post (after deployment) impact assessment.

\subsection{\bf Ex-ante impact assessment}
When impacts are assessed before deployment of a particular system or policy (often using prior knowledge or use cases, empirical data from system testing, and system design details), we talk of ex-ante impact assessment.
Ex-ante impact assessment often predicts the risks and impacts of the proposed system.
Such assessments are used as tools to assess potential environmental, financial, social, and human rights ramifications of systems (both algorithmic and non-algorithmic), projects, and policies, and grant some measure of control and voice to designers or developers, affected population, and authorities to make, induce or enforce changes accordingly. 
~\\\textbf{Environmental impact assessment (EIA):} Both deductive and inductive causal inference have long been used in EIA~\cite{mareddy2017environmental,voegeli2019sustainability}.
While with a deductive approach, a hypothesis about a causal relationship is formed and tested, using an inductive approach, data are collected from observations, and a causal relationship is inferred from the instances. Causal networks have long been used for EIA~\cite{mareddy2017environmental,voegeli2019sustainability} since they bring both network (multiple interaction pathways between environment and various activities) and cause-effect (various activities affecting the elements of the environment) logic to the analysis. Causal networks allow an analysis of impacts through sequences of interactions~\cite{mareddy2017environmental}—--also referred to as \textit{sequence diagrams}~\cite{canter1982environmental}. Causal networks (structural causal models) have helped find indirect impacts on multiple levels~\cite{canter1982environmental}. While initial EIA studies have helped find forms and parameters of relationships between various kinds of activities/developments and environmental elements, they are also  utilized to (ex-ante) estimate the environmental impacts of planned new or upcoming projects, including complex AI systems in the near future~\cite{strubell2019energy}. 
~\\\textbf{Social and Fiscal impact assessment (SIA and FIA):}~\citet{becker2001social} defines SIA as \textit{the process of identifying the future consequences of a current or proposed action which are related to individuals, organizations and social macro-systems}. Similar to the works in EIA, SIA also uses both deductive and inductive methods to discover the causal relationships between actions (e.g. inventory optimization, logistic changes) and consequences (e.g., increase in sales, popularity, overall perception of the product)~\cite{wu2015causality,de2018development}.  
Note that most of the works here use the potential outcomes framework.
Moreover, economists have long been looking for ways to construct counterfactuals and answer causal and policy evaluation questions~\cite{haavelmo1943statistical}. The majority of works on FIA or economic impact assessment try to understand the effects of introducing new economic policies or changing existing ones~\cite{haseeb2019economic}. Such analyses are done in a multi-agent setting with modeling of the preferences and choices of agents along with their ability to infer evaluations and outcomes. Causal and counterfactual economic assessment has been a big part of FIA of many automated systems~\cite{haseeb2019economic}. 

\subsection{\bf Ex-post impact assessment}
Since ex-ante impact assessments are limited to the available prior knowledge and use cases, it is often not possible to identify all possible risks and impacts of a system or policy, which motivates ex-post impact assessment.
When impacts are assessed after deployment (often in real-time using the running record of the system, real-time audits, and evaluations), we call this ex-post assessment.
Ex-post impact assessment often detects the risks and impacts on the go once the system is introduced.
While ex-ante impact assessments often have clear guidelines or metrics for specific types of impacts (e.g., average CO$_2$ emission for environmental impact, total financial cost, opportunity cost and return-on-investment for economic impact, overall positive or negative opinions of the population for social impact), ex-post impact assessments are generally broader since they need to define what constitutes an impact in real-time. Causal inference is used to tackle various categories of risks and impacts of a system or policy in real time.

\noindent\textbf{Temporal and long-term effects:} Many kinds of temporal and long-term effects are seen in real-world systems which operate inside an eco-system of stakeholders. For example, the creation of online filter bubbles and polarization due to personalized consumption-centric information retrieval systems powered by various ML models~\cite{ramaciotti2021auditing,garimella2017long},
    the content homogenization effects observed in online marketplaces as a response to popularity bias in recommender systems~\cite{chaney2018algorithmic}. 
    Causality, along with behavioral modeling, has helped assess such systems in real-time and find out the elements responsible~\cite{sharma2015estimating}.
~\\\textbf{Effects from system failure and system misuse:} Systems are often designed with some desired criteria (e.g., accuracy, fairness, robustness, etc.). If a real-world model deviates from the desired criteria and produces unwanted outputs, \textit{accountability} helps identify the reason behind that failure and take required actions.  
    For example~\citet{gillespie2021impact} studies how the inclusion of a robot as a team member in surgeries increases the complexity and errors that it was supposed to reduce. Making systems and stakeholders accountable for such failures of the system is an integral part of safety.~\citet{ibrahim2021causality} propose a bottom-up causal approach using goal-specific accountability mechanisms. Their mechanisms can help identify the root cause of specific type(s) of events or failures, which can then be used to eliminate the underlying (technical) problem and also to assign blame. 
    Causality has also been very helpful in analyzing (networked or other) system attacks and finding out the root cause and the source host or process from where the attack has originated~\cite{chow2004understanding,liu2018towards}.
~\\\textbf{Strategic risks and effects:} While interpretability and explainability are essential for trustworthiness, one must also consider various risks of using interpretable and explainable decision-making systems in the real world. For example~\citet{shokri2021privacy} analyze connections between model explanations and the leakage of sensitive information in the model’s training set 
    even if a model is used as a black-box; They show that back-propagation-based explanations can leak a significant amount of information about individual training data points, exhibiting a potential conflict between privacy and interpretability. 
Similarly~\citet{tsirtsis2020decisions} show that counterfactual explanations can reveal various details of decision-making systems, thereby making them vulnerable towards strategic behaviors and, therefore, non-robust.     
    Since in real-world settings, individuals (either using the decision-making system or being affected by its decisions) often try to optimize their outcome, their rational strategic behavior can cause issues of privacy and robustness.
    Because of the above strategic risks, recent works~\cite{tsirtsis2020decisions,patro2022fair} suggest modeling real-world strategic settings as games (from applied game theory) and then designing decision-making systems that can mitigate the effects of strategic behavior.
    Essentially, these works~\cite{tsirtsis2020decisions,patro2022fair} 
    try to design decision-making systems that incentivize individuals only to improve a desired quality (to improve individual outcomes) and not do any strategic manipulations.

\subsection{Conclusion}
We discussed the importance of impact assessment and auditing for the safety and accountability of AI systems and the use of causality in different types of ex-ante and ex-post assessment.
However, as~\citet{moss2020governing} outline, there is no pre-existing or universal definition of impact that can be used in safety assessment of algorithmic systems.
Instead, the idea of an impact can vary depending on the context, scale, and application domain.
Adding to that,~\citet{irving2019ai} justifiably argues that figuring out AI safety would need social scientists since it depends on human values and expectations.
Given clear definitions of undesirable effects, causality can be very helpful, particularly in studying the effects of different design elements of algorithmic systems in simulated environments, and also in figuring out which design element(s) are responsible for certain undesirable effects observed in either real or simulated environments.
\section{Causality in Healthcare}
\label{sec:health-care}

In the previous sections, 
we mostly demonstrated the effectiveness of causality-based methods for improving different trustworthy aspects of AI. However, causal methods have proved to be reliable tools in many application domains as well. 
In this section, we look into \emph{healthcare}, where causal methods have been demonstrated to be particularly advantageous. Additionally, we also point out how causal methods could be effective in enhancing the trustworthiness aspects of such applications.

\begin{figure}[t!] 
  \centering  
   \small
    \begin{forest}
      for tree={
        parent anchor=south,
        child anchor=north,
        fit=band,
      }
      [Causality in Healthcare
        [Structural Causal Model 
        [Analysing \\ Diseases \\ \cite{shen2020,Sanchez2022,mani2000}]
        [Drugs \\ Repurposing \\ \cite{belyaeva2021}]
        [Improving \\ Diagnosis \\ 
        \cite{vanAmsterdam2019,richens2020}]
        ]
        [Potential Outcome Framework 
        [Analysing \\ Drugs Outcome \\ \cite{shi2022b,graham2015,ozer2022,friedrich2020,ziff2015,shi2022a}]]
      ]
    \end{forest}
    \caption{Structure of different healthcare applications used with causality.}
    \label{fig:healthcare_tree} 
\end{figure}
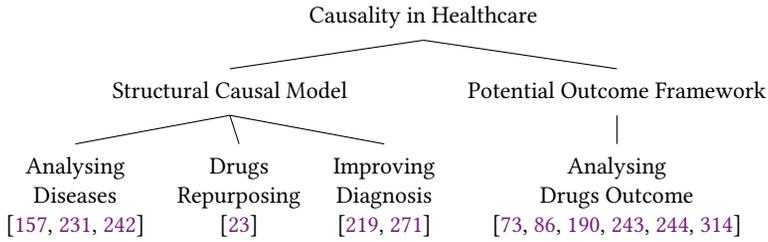

Several existing papers highlight the application of causality approaches in the healthcare domain. Through SCMs, discussed in \textbf{Section~\ref{sec:scm}}, we can better analyze diseases (by revealing the causal relations of the diseases' features), improve the accuracy of automatic diagnosis, or even discover new drugs. Moreover, causal reasoning can also be used to investigate drugs' outcome through the PO framework, discussed in \textbf{Section~\ref{sec:po}}. 
Figure~\ref{fig:healthcare_tree} depicts these different use cases.
In this section, we will survey  existing research  
based on the two broad categories of causal inference, SCMs and PO, while focusing on interpretability, robustness, fairness, and privacy.

\subsection{Causality in Healthcare through SCM framework}
The literature provides many reviews which focus on the usage of SCMs in healthcare and personalized medicine~\cite{zhang2021,Sanchez2022,vlontzos2022}. ~\citet{zhang2021} provided an introduction to causality using different medical examples like lung cancer causal graphs and shed light on the different challenges and issues encountered when dealing with causal inference, such as missing data, biased data, and transferability of models. ~\citet{vlontzos2022} present the benefits of introducing causality and its use in the field of medical imaging. Their survey reviews several applications that incorporate causal discovery and causal inference in medical imaging.
~\citet{Sanchez2022} 
specifically focused on Alzheimer’s disease (AD), a progressive neurodegenerative disorder, to highlight the utility of causal machine learning in precision medicine. 
~\citet{shen2020} used two causal discovery methods to discover the causal relationship utilizing observational data of AD. The authors used the {\it{fast causal inference}} (FCI)~\cite{DBLP:books/daglib/0023012}, a constraint-based algorithm, and the {\it{fast greedy equivalence search}} (FGES)~\cite{ramsey2015}, a score based algorithm. These approaches are then compared and benchmarked with a well-established causal graph.
~\citet{mani2000} tried to identify causal factors of clinical conditions for intensive care unit (ICU) patients using medical discharge reports.
In the above references, we observe that causality methods are mostly used to interpret and explain the outcomes of medical models. Such approaches and ideas make the AI system interpretable by design, which is desirable in healthcare use cases. By providing an explanation and finding causal relations between different factors of a particular disease, different insights are gained, which in turn contribute to a better, trustworthy AI system.

Drug discovery is another research area that benefits from causality.
~\citet{belyaeva2021} show  that causal models can be used to repurpose drugs for new diseases like SARS-CoV-2. The research team integrated transcriptomic, proteomic, and structural data for different diseases. They first used autoencoders to match a drug's signature with a reverse disease signature in the latent space. Using the augmented Steiner tree, the disease interactome is then identified. Lately, they have verified the causal interaction of the drugs with genes by using a causal structure discovery algorithm and building a causal network. 

Apart from interpretability, few studies addressed fairness and robustness issues.
In precision medicine, we aim for a fair system that provides personalized and equitable treatment to each individual without any bias~\cite{chen2021}.  
~\citet{chen2021} gave the example of biased systems in healthcare. An algorithm trained only on USA cancer pathology data may lead to wrong classification, when deployed on data from Turkish cancer patients, due to protocol variations or population shifts (imbalanced data). The authors argued that causality is one of the technologies (others being fairness-aware federated learning, features disentanglement, etc.) that contribute towards a fair algorithm in healthcare by performing causal analysis to identify the bias factors.  We should therefore include causality and analyze the causal structures, which could be discovered or provided by the clinicians to make biased AI algorithms fair in real-world scenarios and healthcare applications. 

Robustness is another aspect of trustworthy AI that is addressed in various causality papers related to healthcare applications.  
It is important to design a system that is robust to  change in the distribution and provides reliable outputs and accurate results.~\citet{vanAmsterdam2019} improved their lung cancer image prediction algorithm by eliminating bias signals. 
By predicting the collider variable (tumor size) and the prognostic factor (tumor heterogeneity), it was then possible to unbias the estimation and make the system more robust.
~\citet{richens2020} improved the accuracy of medical diagnosis through the use of causal machine learning. Their counterfactual algorithm helps them to improve the decision-making process and classify different vignettes correctly based on Bayesian networks that model known relationships between multiple diseases and integrate the causal relationship between different variables.
Such algorithms incorporate the collider variables and also different causal relations in the network architecture design, which makes them in-processing methods that enhance robustness.

\subsection{Causality in Healthcare through the PO framework}
The PO framework is commonly used in the medical field. It provides methods to conduct causal analysis from a statistical perspective, as already introduced in \textbf{Section~\ref{sec:po}}.
In real-world scenarios, observational data are biased (e.g. biased labeling, under-representation, etc.). PO framework methods  provide a way to remove the selection bias in the historical data, thereby leading to a fair system (e.g. the {\textit{propensity score matching}} method: \textbf{Section~\ref{sec:po}}). 

~\citet{shi2022a} 
review different research works which focused on learning causal effects from observational data in the medical domain. These survey articles summarize different methods used to estimate treatment effects.
In the medical field, it is common to use the PO framework to test whether a particular drug is beneficial or harmful. For instance, ~\citet{graham2015} used propensity score matching~\cite{rosenbaum1983} to examine whether Dabigatran or Warfarin increase the risk of death in elderly patients from nonvalvular atrial fibrillation. 
Similarly,~\citet{ozer2022} investigated the benefits of chemotherapy in comparison to only undergoing surgery for patients with resectable gallbladder cancer. By performing propensity score matching~\cite{rosenbaum1983} analysis, it was possible to find out that chemotherapy increases the survival rate.
~\citet{friedrich2020} tried several propensity score-based methods utilizing the PO framework, in addition to other approaches such as g-computation and doubly robust estimators~\cite{hernan2020}. They designed a simulation to compare these different methods. It mimics data from a small non-randomized study on the efficacy of hydroxychloroquine for COVID-19 patients  
~\citet{ziff2015} used PO framework-based analysis (propensity score matching~\cite{rosenbaum1983}) to evaluate the safety and efficacy of the drug digoxin for patients with heart failure.

As discussed in \textbf{Section~\ref{sec:privacy}}, privacy is one of the main foundations required for a trustworthy AI system. An AI system should not allow the identification of specific patients based on the available training datasets. This aspect was one of the major concerns covered in the paper ~\cite{shi2022b}. 
This paper states that the publicly available dataset helps assess a single treatment's efficacy. However, to investigate multiple treatments and correctly estimate the causal effect through observational data, the authors generated a new large synthetic dataset that imitates real-world data distributions and preserves individual patients' privacy.
They reported the $\epsilon${\it{-identifiability}} metric that estimates the probability that an individual is identifiable and ensures that this value remains low after the data generation process.

\subsection{Conclusion}
Work in the healthcare domain encompassing causality mainly considers 
the aspects of interpretability and robustness and touch upon fairness and privacy to some extent. The other tenets of trustworthy AI, like safety and accountability, must be explored further.
Integrating causality into precision medicine is still facing several challenges that need to be addressed. When discovering causal relations (i.e. causal discovery) among different features, ground truth causal graphs are not always available to validate the results~\cite{Holzinger2019}. This implies over-trusting the data, and the discovered relations can be overcome by working closely with experts and clinicians to integrate their knowledge.
Available data in the healthcare domain are commonly unstructured, highly complex, and multimodal~\cite{Sanchez2022}. Thus, there is an urge to develop better decision-making algorithms that not only find correlations in these data but also understand causal relations and perform causal reasoning.

\section{Conclusion}
In this article, we surveyed causal modeling and reasoning tools for enhancing the trustworthy aspects of AI models, which include interpretability, fairness, robustness, privacy, safety, and accountability. While in recent years, the community has witnessed an unprecedented surge of research in this context, important facets still remain unexplored.
We expect significant advancements in the coming years and hope this survey will act as an important resource to the community and at the same time guide future research connecting trustworthy AI and causality. 
\begin{acks}
This work has received funding from the European Union’s Horizon 2020 research and innovation programme under Marie Sklodowska-Curie Action ``NoBIAS - Artificial Intelligence without Bias'' (grant agreement number 860630) and Network of Excellence ``TAILOR - A Network for Trustworthy Artificial Intelligence'' (grant agreement number 952215), the Lower Saxony Ministry of Science and Culture under grant number ZN3492 within the Lower Saxony ``Vorab'' of the Volkswagen Foundation and supported by the Center for Digital Innovations (ZDIN), and the Federal Ministry of Education and Research (BMBF), Germany under the project ``LeibnizKILabor'' with grant No. 01DD20003 and from Volkswagen Foundation and the Ministry for Science and Culture of Lower Saxony, Germany (MWK) under the "Understanding Cochlear Implant Outcome Variability using Big Data and Machine Learning Approaches" (grant no. ZN3429) project.
\end{acks}

\bibliographystyle{ACM-Reference-Format-num}
\bibliography{References}

\newpage
\appendix
\section*{A Review of the Role of Causality in Developing Trustworthy AI Systems –- Datasets and Packages}

As a result of our literature review on causality-based solutions for Trustworthy AI, a need for an extensive overview of relevant datasets and packages was observed. To make causal machine learning (ML) more accessible and to facilitate comparisons to non-causal methods, we created a curated list of datasets used for recent Causal ML publications. This appendix also includes an overview of useful causal and non-causal tools and packages to assess different trustworthy aspects of ML models (interpretability, fairness, robustness, privacy, and safety). We also provide a similar overview for the healthcare domain. Each aspect has its dedicated section that is structured as follows:

\begin{enumerate}
    \item An overview of \textbf{publicly available real-world datasets} used in cited publications of this survey
    \item Some \textbf{benchmarks and packages for Causal Machine Learning} that researchers could utilize
    \item A number of \textbf{well-established tools}, that allow for a better comparison to non-causal machine learning
\end{enumerate}
We want to clarify that this section does not (and cannot) aim for completeness. Instead, we want to provide researchers interested in working on a selection of aspects of Trustworthy AI with a concise overview of exciting avenues for experimenting with causal machine learning. The resources are hyperlinked and sorted based on when their associated causal papers first appear in the corresponding subsections (e.g., datasets used in pre-processing papers will appear first). We highly encourage readers to seek additional reading material, such as Chapter 9 of~\cite{kaddour_2022_causalmlsurvey} or the two Github repositories for datasets\footnote{\url{https://github.com/rguo12/awesome-causality-data}} and algorithms\footnote{\url{https://github.com/rguo12/awesome-causality-algorithms}} resulting from~\cite{guo2020survey}.

\section{Interpretability}
\label{sec:appendix_interpretability}

\subsection{Datasets Used by Cited Publications}

\begin{itemize}

    \item \textbf{\href{https://github.com/causal-disentanglement/CANDLE}{CANDLE}}~\citep{reddy:2022:AAAI:candle_disentangled_dataset}: A dataset of realistic images of objects in a specific scene generated based on observed and unobserved confounders (object, size, color, rotation, light, and scene). As each of the 12546 images is annotated with the ground-truth information of the six generating factors, it is possible to emulate interventions on image features. $\xrightarrow[]{}$~\textbf{Used by}:~\cite{reddy:2022:AAAI:candle_disentangled_dataset}

    \item \textbf{\href{https://msnews.github.io/}{MIND}}~\cite{wu2020mind}: A news recommendation dataset built upon user click logs of Microsoft News. It contains 15 million impression logs describing the click behavior of more than 1 Million users across over 160k English news articles. Each news article entry contains its title, category, abstract, and body. Each log entry is made up of the users' click events, non-clicked events, and historical news click behaviors prior to this impression. $\xrightarrow[]{}$~\textbf{Used by}:~\cite{si:2022:www:causalRecSearch}
    
    \item \textbf{\href{https://grouplens.org/datasets/movielens/}{MovieLens}}~\cite{MovieLens2015}: A group of datasets containing movie ratings between 0 and 5 (with 0.5 increments) collected from the MovieLens website. Movies are described through their title, genre, and relevance scores of tags (e.g., romantic or funny). GroupLens Research constantly releases new up-to-date MovieLens databases in different sizes. $\xrightarrow[]{}$~\textbf{Used by}:~\cite{zheng:2021:www:disentangeluserinterestConformity}
    
    \item \textbf{\href{https://www.kaggle.com/datasets/netflix-inc/netflix-prize-data}{Netflix Prize}}~\cite{bennett2007netflix}: A movie rating dataset consisting of about 100 Million ratings for 17,770 movies given by 480,189 users. Ratings consists of four entries: user, movie title, date of grade, and a grade ranging from 1 to 5. Users and movies are represented with integer IDs. $\xrightarrow[]{}$~\textbf{Used by}:~\cite{zheng:2021:www:disentangeluserinterestConformity}

    \item \textbf{\href{https://www.statmt.org/wmt14/}{WMT 14}}~\cite{bojar2014findings}: WMT\footnote{\url{https://machinetranslate.org/wmt}} is a yearly workshop in which researchers develop machine translation models for several different tasks. WMT14 was created for the event in 2014 and included a translation, a quality estimation, a metrics, and a medical translation task. Each category comprises different subtasks (e.g., translating between two specific languages).  $\xrightarrow[]{}$~\textbf{Used by}:~\cite{alvarez2017causal}

    \item \textbf{\href{https://opus.nlpl.eu/OpenSubtitles-v2018.php}{OpenSubtitles}}~\cite{lison2016opensubtitles2016}: A text corpus comprising over 2.6 billion sentences from movie dialogues. The data stem from pre-processing 3,735,070 files from the online database \textit{OpenSubtitles.org}\footnote{\url{https://www.opensubtitles.org/}}. This corpus covers dialogues from ca. 2.8 million movies in 62 languages. $\xrightarrow[]{}$~\textbf{Used by}:~\cite{alvarez2017causal}
    
    \item \textbf{\href{https://github.com/facebookresearch/LAMA}{LAMA}}~\cite{petroni2019language}: A probe designed to examine the factual and commonsense knowledge in pretrained language models. It is built upon four different, prominent corpora of facts that cover a wide range of knowledge types. $\xrightarrow[]{}$~\textbf{Used by}:~\cite{cao:2022:acl:promtLMBiasCausal}
    
    \item \textbf{\href{https://github.com/commaai/research}{Comma.ai Driving Dataset}}~\cite{santana2016learning}: A video dataset made up of 11 video clips of variable size capturing the windshield view of an Acura ILX 2016. The driving data contains 7.25 hours of footage, which was mostly recorded on highways. Each video is accompanied by measurements such as the car's speed, acceleration, or steering angle. $\xrightarrow[]{}$~\textbf{Used by}:~\cite{Kim_2017_ICCV}
    
    \item \textbf{\href{https://github.com/udacity/self-driving-car}{Udacity Driving Dataset}}~\cite{UdacityDriving}: A driving video dataset developed for the Udacity \textit{Self-Driving Car Nanodegree Program}\footnote{\url{https://udacity.com/self-driving-car}}. The GitHub repository contains two annotated datasets in which computer vision systems have to label objects, such as cars or pedestrians, within driving footage. $\xrightarrow[]{}$~\textbf{Used by}:~\cite{Kim_2017_ICCV}
    
    \item \textbf{\href{https://github.com/hadyelsahar/RE-NLG-Dataset}{T-REx}}~\cite{elsahar2018trex}: A dataset of large-scale alignments between Wikipedia abstracts and Wikidata triples. Such triples encode semantic information in the form of subject-predicate-object relationships. T-REx consists of 11 million triples with 3.09 million Wikipedia abstracts (6.2 million sentences). $\xrightarrow[]{}$~\textbf{Used by}:~\cite{li:2022:acl:PTCaptureFactsCausal}
    
    \item \textbf{\href{http://yann.lecun.com/exdb/mnist/}{MNIST}}~\cite{lecun1998gradient}: An extraordinarily well-known and widely used image dataset comprising 28 $\times$ 28 grayscale images of handwritten digits. It contains 60,000 training and 10,000 test samples. $\xrightarrow[]{}$~\textbf{Used by}:~\cite{Schwab2019}
    
    \item \textbf{\href{https://www.image-net.org/about.php}{ImageNet}}~\cite{ImageNet2009}: Another well-known, more sophisticated image dataset containing more than 14 million images. The images depict more than 20,000 \textit{synsets} (i.e., concepts "possibly described by multiple words or word phrases"\footnote{\url{https://www.image-net.org/about.php}}). $\xrightarrow[]{}$~\textbf{Used by}:~\cite{Schwab2019}
    
    \item \textbf{\href{https://archive.ics.uci.edu/ml/datasets/adult}{Adult (Census Income)}}~\cite{kohavi1996ADULTS, dheeru2017uci}: A tabular dataset containing anonymized data from the 1994 Census bureau database.\footnote{\url{http://www.census.gov/en.html}} Classifiers try to predict whether a given person will earn over or under 50,000 USD worth of salary. Each person is described via 15 features (including their id), e.g., gender, education, and occupation. $\xrightarrow[]{}$~\textbf{Used by}:~\cite{NEURIPS2020_0d770c49, Mahajan2019}
    
    \item \textbf{\href{https://archive.ics.uci.edu/ml/datasets/human+activity+recognition+using+smartphones}{Human Activity Recognition}}~\cite{anguita2013public}: This dataset contains smartphone-recorded sensor data from 30 subjects performing \textit{Activities of Daily Living}. The database differentiates between the activities walking (upstairs, downstairs, or on the same level), sitting, standing, and laying. $\xrightarrow[]{}$~\textbf{Used by}:~\cite{Janzing2020}
    
    \item \textbf{\href{https://www.yelp.com/dataset}{Yelp}}~\cite{YelpDataset}: A dataset of almost 7 million Yelp user reviews of around 150k businesses across 11 cities in the US and Canada. Review entries contain not only their associated text and an integer star rating between 1 and 5 but also additional information like the amount of \textit{useful}, \textit{funny}, and \textit{cool} votes for the review. $\xrightarrow[]{}$~\textbf{Used by}:~\cite{Tan2021}
    
    \item \textbf{\href{https://cseweb.ucsd.edu/~jmcauley/datasets/amazon_v2/}{Amazon (Product) Data}}~\cite{ni2019justifying}: An extensive dataset of 233.1 million Amazon reviews between May 1996 and October 2018. The data include not only information about the review itself and product metadata (e.g., descriptions, price, product size, or package type) but also \textit{also bought} and \textit{also viewed} links. $\xrightarrow[]{}$~\textbf{Used by}:~\cite{Tan2021}

    \item \textbf{\href{https://github.com/divyat09/cf-feasibility}{Sangiovese Grapes}}~\cite{magrini2017conditional}: A conditional linear Bayesian network that captures the effects of different canopy management techniques on the quality of Sangiovese grapes. Based on a two-year study of Tuscan Sangiovese grapes, the authors created a network with 14 features (13 of which are continuous variables). The data used for experiments in \cite{Mahajan2019} are linked in their repository (see the link behind the term ``Sangiovese Grapes''). $\xrightarrow[]{}$~\textbf{Used by}:~\cite{Mahajan2019}
    
    \item \textbf{\href{https://blog.salesforceairesearch.com/the-wikitext-long-term-dependency-language-modeling-dataset/}{WikiText-2}}~\cite{merity2016pointer}: An NLP benchmark containing over 100 million tokens extracted from verified Good and Featured articles on Wikipedia. Contrary to previous token collections, however, WikiText-2 is more extensive and comprises more realistic tokens (e.g., lower-case tokens). $\xrightarrow[]{}$~\textbf{Used by}:~\cite{jeoung:2022:arxiv:DebiasCausalMediation}
    
    \item \textbf{\href{https://www.kaggle.com/c/jigsaw-unintended-bias-in-toxicity-classification}{Jigsaw Toxicity Detection}}~\cite{Jigsaw2019}: A dataset of comments made across around 50 English-language news sites built to analyze unintended bias in toxicity classification within a Kaggle competition organized by Jigsaw and Google. Each comment in the training set comes with a human-annotated toxicity label (e.g., obscene or threat) and labels for mentioned identities (e.g., gender, ethnicity, sexuality, or religion) in the comment. $\xrightarrow[]{}$~\textbf{Used by}:~\cite{jeoung:2022:arxiv:DebiasCausalMediation}
    
    \item \textbf{\href{https://nlp.stanford.edu/robvoigt/rtgender/}{RTGender}}~\cite{voigt2018rtgender}: A collection of comments made on online content across different platforms such as Facebook or Reddit. Each post and comment is annotated with the gender of the author in order to analyze gender bias in social media. $\xrightarrow[]{}$~\textbf{Used by}:~\cite{jeoung:2022:arxiv:DebiasCausalMediation}
    
    \item \textbf{\href{https://github.com/nyu-mll/crows-pairs}{CrowS-Pairs}}~\cite{nangia2020crows}: A benchmark designed to investigate the social bias of NLP models. Each entry consists of two sentences: one representing a stereotypical statement for a given bias type (e.g., religion or nationality) and an anti-stereotypical version of the statement, where the described group/identity was substituted. $\xrightarrow[]{}$~\textbf{Used by}:~\cite{jeoung:2022:arxiv:DebiasCausalMediation}

    \item \textbf{\href{https://github.com/sebastianGehrmann/CausalMediationAnalysis}{Professions}}~\cite{DBLP:conf/nips/VigGBQNSS20}: A set of templates (originating from \cite{lu2020gender}) that were augmented with professions from \cite{bolukbasi2016man}. Each sentence template follows the pattern “The [occupation] [verb] because”, and each profession has a crowdsourced rating that describes its definitionality and stereotypicality. $\xrightarrow[]{}$~\textbf{Used by}:~\cite{DBLP:conf/nips/VigGBQNSS20}
    
    \item \textbf{\href{https://uclanlp.github.io/corefBias/overview}{WinoBias}}~\cite{WinoBias2018}: A collection of 3,160 WinoCoRef style sentences created to estimate gender bias within NLP models. Sentences come in pairs that only differ by the gender of one pronoun, with each sentence describing an interaction between two people with different occupations. $\xrightarrow[]{}$~\textbf{Used by}:~\cite{DBLP:conf/nips/VigGBQNSS20}
    
    \item \textbf{\href{https://github.com/rudinger/winogender-schemas}{Winogender Schemas}}~\cite{Winogender2018}: A Winograd-style collection of templates that generate pairs of sentences that only differ by the gender of one pronoun. Researchers can generate 720 different sentences by defining the building blocks \textit{occupation}, \textit{participant}, and \textit{pronoun} as a benchmark for gender bias detection. $\xrightarrow[]{}$~\textbf{Used by}:~\cite{DBLP:conf/nips/VigGBQNSS20}
    
    \item \textbf{\href{https://universaldependencies.org/en/}{English UD Treebank}}~\cite{mcdonald2013universal}: The English UD Treebanks represents a subset of a data collection containing uniformly analyzed sentences across six different languages. The English treebank consists of 43,948 sentences and  1,046,829 tokens. $\xrightarrow[]{}$~\textbf{Used by}:~\cite{DBLP:journals/tacl/ElazarRJG21}
    
    \item \textbf{\href{https://github.com/uclanlp/gn_glove}{Gender-Neutral GloVe Word Embeddings}}~\cite{zhao2018learning}: This variant of GloVe produces gender-neutral word embeddings by maintaining all gender-related information exclusively in specific dimensions of word vectors. The resulting word embeddings can be a starting point for more unbiased NLP. $\xrightarrow[]{}$~\textbf{Used by}:~\cite{ravfogel:2020:acl:INLP, ravfogel:2022:acl:RLACE}
    
    \item \textbf{\href{https://github.com/Microsoft/biosbias}{Biographies}}~\cite{de2019bias}: A collection of 397,340 online biographies covering 28 occupations (e.g., professors, physicians, or rappers). Each biography is stored as a dictionary containing the title, the (binary) gender, the length of the first sentence, and the entire text of the biography. $\xrightarrow[]{}$~\textbf{Used by}:~\cite{ravfogel:2020:acl:INLP, ravfogel:2022:acl:RLACE}
    
    \item \textbf{\href{http://slanglab.cs.umass.edu/TwitterAAE/}{TwitterAAE corpus}}~\cite{blodgett2016demographic}: A collection of 59.2 million tweets sent out by 2.8 million users from the US in 2013. Each tweet is annotated with a vector describing the ``likely demographics of the author and the neighborhood they live in.''~\cite{blodgett2016demographic} These demographic approximations of users were built upon US census data. $\xrightarrow[]{}$~\textbf{Used by}:~\cite{ravfogel:2020:acl:INLP}
    
    \item \textbf{\href{https://mmlab.ie.cuhk.edu.hk/projects/CelebA.html}{CelebA}}~\cite{liu2015faceattributes}: A face image dataset containing 202,599 images of size 178 $\times$ 218 from 10,177 unique celebrities. Each image is annotated with 40 binary facial attributes (e.g., \textit{Is this person smiling?}) and five landmark positions describing the 2D position of the eyes, the nose, and the mouth (split into \textit{left} and \textit{right} side of the mouth). $\xrightarrow[]{}$~\textbf{Used by}:~\cite{ravfogel:2022:acl:RLACE}
    
\end{itemize}

\subsection{Interesting Causal Tools}

\begin{itemize}
    \item \textbf{\href{https://github.com/kuiy/CausalFS}{CausalFS}}~\cite{Yu2020}: An open-source package for C++ that contains 28 local causal structure learning methods for feature selection. It is specifically designed to facilitate the development and benchmarking of new causal feature selection techniques.  
    
    \item \textbf{\href{https://github.com/CEBaBing/CEBaB}{CEBaB}}~\cite{abraham2022cebab}: A recently designed benchmark to estimate and compare the quality of concept-based explanation for NLP. CEBaB includes a set of restaurant reviews accompanied by human-generated counterfactuals, which enables researchers to investigate the model's ability to assess causal concept effects.
    
    \item \textbf{\href{https://github.com/amirfeder/CausaLM}{CausaLM Datasets}}~\cite{feder:2021:coli:CausaLM}: As part of the analysis of \textit{CausaLM}, the authors developed four NLP datasets for evaluating causal explanations. These datasets represent real-world applications of ML that come with ground-truth information.
    
    \item \textbf{Competing with Causal Toolboxes}: Several causal tools like \textit{\href{https://github.com/DataCanvasIO/YLearn}{YLearn}}~\cite{YLearn2022}, \textit{\href{https://github.com/py-why/dowhy}{DoWhy}}~\cite{DoWhy_2019_Microsoft}, \textit{\href{https://github.com/uber/causalml}{CausalML}}~\cite{Chen2020_CausalML_Uber}, or \textit{\href{https://github.com/microsoft/EconML}{EconML}}~\cite{EconML_2019_Microsoft} introduce an entire causal inference pipeline with their own interpreter module. Comparing newly developed interpretation techniques with such packages could be very insightful.
    
\end{itemize}

\subsection{Prominent Non-Causal Tools}

\begin{itemize}

    \item \textbf{\href{https://github.com/marcotcr/lime}{LIME}}~\cite{ribeiro2016LIME}: A very prominent Python package that allows researchers to explain individual predictions of image, text, and tabular data classifiers. Applicable to any black-box classifier that implements a function that outputs class probabilities given raw text or a NumPy array.
    
    \item \textbf{\href{https://github.com/google-research/google-research/tree/master/interpretability_benchmark}{ROAR}}~\cite{hooker2019ROAR}: A benchmark method that evaluates interpretability approaches based on how well they quantify feature importance. The technique was used to assess model explanations of image classifiers over multiple datasets.
    
    \item \textbf{\href{https://github.com/slundberg/shap}{SHAP}}~\cite{lundberg2017unified}: Another well-known interpretability package which is based on game theory. Although compatible with any ML model, SHAP comes with a C++-based algorithm for tree ensemble algorithms such as XGBoost.
    
    \item \textbf{\href{https://github.com/interpretml/interpret}{InterpretML}}~\cite{nori2019interpretml}: An open-source package developed by Microsoft that includes multiple state-of-the-art methods for model interpretability. It also allows users to train an \textit{Explainable Boosting Machine} (EBM) - a model that provides exact explanations and performs as well as random forests and gradient-boosted trees. 
    
\end{itemize}

\section{Fairness}
\label{sec:appendix_fairness}

\subsection{Datasets Used by Cited Publications}
\begin{itemize}
    \item \textbf{\href{https://archive.ics.uci.edu/ml/datasets/adult}{Adult (Census Income)}}~\cite{kohavi1996ADULTS, dheeru2017uci}: A tabular dataset containing anonymized data from the 1994 Census bureau database.\footnote{\url{http://www.census.gov/en.html}} Classifiers try to predict whether a given person will earn over or under 50,000 USD worth of salary. Each person is described via 15 features (including their id), e.g., gender, education, and occupation. $\xrightarrow[]{}$~\textbf{Used by}:~\cite{xu2019achieving, zhang2017causal, zhang2018causal, galhotra2022causal, nabi2018fair, pan2021explaining, yan2020silva, salimi2019interventional, wu2019counterfactual} 
    
    \item \textbf{\href{https://github.com/propublica/compas-analysis/}{COMPAS Recidivism Risk}}~\cite{angwin2016COMPAS}: A set of criminological datasets published by ProPublica to evaluate the bias of COMPAS - an algorithm used to assess the likelihood of criminal defendants reoffending. All COMPAS-related datasets include data from over 10,000 defendants, each being described via 52 features (e.g., age, sex, race) and with a label indicating whether they were rearrested within two years. $\xrightarrow[]{}$~\textbf{Used by}:~\cite{galhotra2022causal, chiappa2018causal, nabi2019learning, nabi2018fair, salimi2019interventional, mishler2021fairness}

    \item \textbf{\href{https://community.fico.com/s/explainable-machine-learning-challenge?tabset-3158a=2&tabset-158d9=3}{FICO Credit Risk}}~\cite{board2007FICO}: In this dataset, ML models have to predict whether or not credit applicants will at least once be more than 90 days due with their payment within a two-year timespan. It includes anonymized information about HELOC applicants described through 23 features (e.g., months since the most recent delinquency or number of inquiries in last 6 months)~\cite{chen2018interpretable} $\xrightarrow[]{}$~\textbf{Used by}:~\cite{creager2020causal}
    
    \item \textbf{\href{https://archive.ics.uci.edu/ml/datasets/statlog+(german+credit+data)}{German Credit Risk}}~\cite{dheeru2017uci}: A collection of data from 1,000 anonymized German bank account holders that applied for a credit. Based on the 20 features of the applicant and their application (e.g., credit history, purpose of credit, or employment status), models need to estimate the risk of giving the person a credit and categorize them as either good or bad credit recipients.. $\xrightarrow[]{}$~\textbf{Used by}:~\cite{galhotra2022causal}
    
    \item \textbf{\href{https://meps.ahrq.gov/mepsweb/data_stats/data_overview.jsp}{Medical Expenditure (MEPS})}~\cite{MEPS}: A collection of large-scale surveys of US citizens, their medical providers, and employers. It includes information like race, gender, and the ICD-10 code of the diagnosis of a patient. The given information can be used to predict the total number of patients' hospital visits. $\xrightarrow[]{}$~\textbf{Used by}:~\cite{galhotra2022causal}
    
    \item \textbf{\href{https://mimic.mit.edu/docs/gettingstarted/}{MIMIC III}}~\cite{johnson2016mimic}: A dataset of anonymized clinical records of the Beth Israel Deaconess Medical Center in Boston, Massachusetts. Records contain information like ICD-9 codes for diagnoses and medical procedures, vital signs, medication, or even imaging data. The dataset includes records from 38,597 distinct adult patients. $\xrightarrow[]{}$~\textbf{Used by}:~\cite{singh2021fairness}
    
    \item \textbf{\href{https://grouplens.org/datasets/movielens/}{MovieLens}}~\cite{MovieLens2015}: A group of datasets containing movie ratings between 0 and 5 (with 0.5 increments) collected from the MovieLens website. Movies are described through their title, genre, and relevance scores of tags (e.g., romantic or funny). GroupLens Research constantly releases new up-to-date MovieLens databases in different sizes. $\xrightarrow[]{}$~\textbf{Used by}:~\cite{li2021towards}
    
    \item \textbf{\href{https://www.kaggle.com/mrmorj/insurance-recommendation}{Zimnat Insurance Recommendation}}~\cite{Insurance2020}: A data collection of almost 40,000 Zimnat (a Zimbabwean insurance provider) customers. The data contain personal information (e.g., marital status or occupation) and the insurance products that the customers own. In inference time, models must predict which product was artificially removed based on customer information. $\xrightarrow[]{}$~\textbf{Used by}:~\cite{li2021towards}

    \item \textbf{\href{https://ocrdata.ed.gov/resources/downloaddatafile}{Civil Rights Data Collection (CRDC)}}~\cite{CRDC_2023}: This is an online collection of education-related data. Since 1968, the U.S. Department of Education’s Office for Civil Rights (OCR) biennially collects data from U.S. public primary and secondary schools. The dataset includes information such as race distribution, the percentage of students who take college entrance exams, or whether specific courses (e.g., Calculus) are offered. $\xrightarrow[]{}$~\textbf{Used by}:~\cite{kusner2019making}
    
    \item \textbf{\href{https://discovery.cs.illinois.edu/dataset/berkeley/}{Berkeley}}~\cite{bickel1975sex}: A simple gender bias dataset published back in 1975 containing information on all 12,763 applicants to the University of California, Berkeley graduate programs in Fall 1973. Each candidate entry consists of the candidate's major, gender, year of application (always 1973), and whether they were accepted. $\xrightarrow[]{}$~\textbf{Used by}:~\cite{yan2020silva}
    
\end{itemize}

\subsection{Interesting Causal Tools}
\begin{itemize}
    \item \textbf{Collection of Annotated Datasets}~\cite{le2022survey}: As part of a survey that provides a thorough overview of commonly used datasets for evaluating the fairness of ML,~\citeauthor{le2022survey} generated Bayesian Networks encompassing the relationships of attributes for each dataset. This information could be used as a reference point for potential causal annotations of fairness-related datasets.
    
    \item \textbf{\href{https://github.com/mrtzh/whynot}{WhyNot}}~\cite{miller2020whynot}: A Python package that provides researchers with many simulation environments for analyzing causal inference and decision-making in a dynamic setting. It allows benchmarking of multiple decision-making systems on 13 different simulators. Crucially for this section, WhyNot also enables comparisons based on other evaluation criteria, such as the fairness of the decision-making.
    
    \item \textbf{\href{https://github.com/huawei-noah/trustworthyAI/tree/master/gcastle}{gCastle}}~\cite{zhang2021gcastle}: An end-to-end causal structure learning toolbox that is equipped with 19 techniques for Causal Discovery. It also assists users in data generation and evaluating learned structures. Having a firm understanding of the causal structure is crucial for fairness-related research.
    
    \item \textbf{\href{https://github.com/felixleopoldo/benchpress}{Benchpress}}~\cite{rios2021benchpress}: A benchmark for causal structure learning allowing users to compare their causal discovery methods with over 40 variations of state-of-the-art algorithms. The plethora of available techniques in this single tool could facilitate research into fair ML through causality.
    
    \item \textbf{\href{https://github.com/uber/causalml}{CausalML}}~\cite{Chen2020_CausalML_Uber}: The Python package enables users to analyze the Conditional Average Treatment Effect (CATE) or Individual Treatment Effect (ITE) observable in experimental data. The package includes tree-based algorithms, meta-learner algorithms, instrumental variable algorithms, and neural-network-based algorithms. Fair-ML researchers could use the provided methods to investigate the causal effect of sensitive attributes on the predicted outcome. 
    
\end{itemize}

\subsection{Prominent Non-Causal Tools}
\begin{itemize}

    \item \textbf{\href{https://github.com/Trusted-AI/AIF360}{AI Fairness 360}}~\cite{aif360-oct-2018}: An open-source library (compatible with both Python and R) that allows researchers to measure and mitigate possible bias within their models/algorithms. It includes six real-world datasets, five fairness metrics, and 15 bias mitigation algorithms.

    \item \textbf{\href{https://github.com/fairlearn/fairlearn}{Fairlearn}}~\cite{bird2020fairlearn}: A Python package developed by Microsoft, which is part of the Responsible AI toolbox\footnote{\url{https://github.com/microsoft/responsible-ai-toolbox}}. It contains various fairness metrics, six unfairness-mitigating algorithms, and five datasets.

    \item \textbf{\href{https://github.com/dssg/aequitas}{Aequitas}}~\cite{2018aequitas}: An open-source auditing tool designed to assess the bias of algorithmic decision-making systems. It provides utility for evaluating the bias of decision-making outcomes and enables users to assess the bias of actions taken directly. 
    
    \item \textbf{\href{https://github.com/google/ml-fairness-gym}{ML-Fairness-Gym}}~\cite{fairness_gym}: A third-party extension of the OpenAI gym designed to analyze bias within RL agents. Although not built upon real-world data, the simulations developed for this benchmark can lead to insights applicable to the real world. It comes with four simulation environments.

\end{itemize}

\section{Robustness}
\label{sec:appendix_robustness}

\subsection{Datasets Used by Cited Publications}
\begin{itemize}
    \item \textbf{\href{https://github.com/ghif/mtae}{Rotated MNIST}}~\cite{ghifary2015domain}: The dataset consists of MNIST images with each domain containing images rotated by a particular angle $\{0^{\circ}, 15^{\circ}, 30^{\circ}, 45^{\circ}, 60^{\circ}, 75^{\circ}\}$ $\xrightarrow[]{}$~\textbf{Used by}:~\cite{Ilse_2021_simulatinginterventions}
    
    \item \textbf{\href{https://github.com/facebookresearch/InvariantRiskMinimization}{ColoredMNIST}}~\cite{arjovsky2019invariant}: The dataset consists of input images with digits 0-4 colored red and labelled 0 while digits 5-9 are colored green representing the two domains. $\xrightarrow[]{}$~\textbf{Used by}:~\cite{Ilse_2021_simulatinginterventions, arjovsky2019invariant, lu2021invariant}
    
    \item \textbf{\href{https://github.com/facebookresearch/DomainBed}{PACS}}~\cite{li2017deeper}: An image classification dataset categorized into 10 classes that are scattered across four different domains, each having a distinct trait: photograph, art, cartoon and sketch. $\xrightarrow[]{}$~\textbf{Used by}:~\cite{Ilse_2021_simulatinginterventions}
    
    \item \textbf{\href{https://cseweb.ucsd.edu/~jmcauley/datasets/amazon_v2/}{Amazon (Product) Data}}~\cite{ni2019justifying}: An extensive dataset of 233.1 million Amazon reviews between May 1996 and October 2018. The data include not only information about the review itself and product metadata (e.g., descriptions, price, product size, or package type) but also \textit{also bought} and \textit{also viewed} links. $\xrightarrow[]{}$~\textbf{Used by}\footnote{\cite{wang2021enhancing} used a subset of Kindle Book reviews from an older version of this dataset}:~\cite{kaushik2020explaining, wang2021enhancing}
    
    \item \textbf{\href{https://alt.qcri.org/semeval2017/task4/}{SemEval-2017 Task 4}}~\cite{rosenthal-etal-2017-semeval}: SemEval\footnote{\url{https://semeval.github.io/}} is a yearly NLP workshop where participants compete on different sentiment analysis tasks. Each workshop comes with its own set of tasks to solve. In SemEval-2017 Task 4, NLP models compete on sentiment analysis tasks on English and Arabic Twitter data. $\xrightarrow[]{}$~\textbf{Used by}:~\cite{kaushik2020explaining}
    
    \item \textbf{\href{https://www.yelp.com/dataset}{Yelp}}~\cite{YelpDataset}: A dataset of almost 7 million Yelp user reviews of around 150k businesses across 11 cities in the US and Canada. Review entries contain not only their associated text and an integer star rating between 1 and 5 but also additional information like the amount of \textit{useful}, \textit{funny}, and \textit{cool} votes for the review. $\xrightarrow[]{}$~\textbf{Used by}:~\cite{kaushik2020explaining}
    
    \item \textbf{\href{https://github.com/acmi-lab/counterfactually-augmented-data}{IMDb extension}}~\cite{kaushik2019learning}: A set of 2440 IMDb reviews, where a human-annotated counterfactual example accompanies each review. The human annotators were found through Amazon’s crowdsourcing platform \textit{Mechanical Turk}\footnote{\url{https://www.mturk.com/}}. The dataset is designed to assess the performance of sentiment analysis and natural language inference models. $\xrightarrow[]{}$~\textbf{Used by}:~\cite{kaushik2020explaining, teney2020learning, wang2021enhancing}

    \item \textbf{\href{https://github.com/acmi-lab/counterfactually-augmented-data}{SNLI extension}}~\cite{kaushik2019learning}: The original SNLI dataset~\cite{bowman2015large} is a text dataset developed to evaluate natural language inference (NLI) models. Models must decide whether a given hypothesis is contradictory to, entailed by, or neutral to the given premise. ~\citet{kaushik2019learning} extended this dataset via humanly-manufactured counterfactual examples. $\xrightarrow[]{}$~\textbf{Used by}:~\cite{kaushik2020explaining, teney2020learning}

    \item \textbf{Parkinson’s voice data}~\cite{little2019causal}: A set of extracted features from audio samples (i.e., sustained phonations) of patients with Parkinson's disease and people from healthy control groups. This dataset combines data from three different and independent labs from the \href{https://archive.ics.uci.edu/ml/datasets/Parkinsons}{US}, \href{http://archive.ics.uci.edu/ml/datasets/Parkinson's+Disease+Classification}{Turkey}, and \href{https://archive.ics.uci.edu/ml/datasets/Parkinson+Dataset+with+replicated+acoustic+features+}{Spain}. The classification task is to detect patients with Parkinson's disease. $\xrightarrow[]{}$~\textbf{Used by}:~\cite{little2019causal}

    \item \textbf{\href{http://ai.bu.edu/M3SDA/}{DomainNet}}~\cite{DomainNet_2019}: An unsupervised domain adaptation image dataset containing six domains (referring to the ``style'' of the image, e.g., sketch, quick drawing or real image) and about ca. 600k images distributed among 345 categories.
    
    \item \textbf{\href{https://www.image-net.org/about.php}{ImageNet}}~\cite{ImageNet2009}: Another well-known, more sophisticated image dataset containing more than 14 million images. The images depict more than 20,000 \textit{synsets} (i.e., concepts "possibly described by multiple words or word phrases"\footnote{\url{https://www.image-net.org/about.php}}). $\xrightarrow[]{}$~\textbf{Used by}:~\cite{Mao_2021_generative, Mitrovic_2020_relic}
    
    \item \textbf{\href{https://github.com/hendrycks/robustness}{ImageNet-C}}~\cite{hendrycks2019benchmarking}: This dataset tests the model's robustness by applying corruptions to validation images of ImageNet. Each of the 15 corruption types (e.g., gaussian noise, snow, motion blur, or contrast) comes with five levels of corruption intensity. $\xrightarrow[]{}$~\textbf{Used by}:~\cite{Mao_2021_generative}
    
    \item \textbf{\href{https://github.com/modestyachts/ImageNetV2}{ImageNet-V2}}~\cite{recht2019imagenet}: A new test set for ImageNet designed to assess the model's generalization ability. Despite closely following the original dataset creation process, models trained on the original ImageNet demonstrate worse performance on ImageNet-V2. ImageNet models with better generalization should perform stably on both variants. $\xrightarrow[]{}$~\textbf{Used by}:~\cite{Mao_2021_generative}
    
    \item \textbf{\href{https://objectnet.dev/}{ObjectNet}}~\cite{barbu2019objectnet}: An image dataset designed to demonstrate the transfer learning ability of ImageNet models. Due to this, ObjectNet provides no training set. Instead, all 50,000 images of ObjectNet combine into a single test set. Each image depicts an object with random backgrounds, viewpoints, and rotations of the object. $\xrightarrow[]{}$~\textbf{Used by}:~\cite{Mao_2021_generative}
    
    \item \textbf{\href{https://www.tensorflow.org/datasets/catalog/imagenet2012}{ImageNet ILSVRC-2012}}~\cite{russakovsky2015imagenet}: The dataset used for the ImageNet Large Scale Visual Recognition Challenge 2012 (ILSVRC2012). The 1.5 million images depict objects from 1,000 different synsets. $\xrightarrow[]{}$~\textbf{Used by}:~\cite{Mitrovic_2020_relic}

    \item \textbf{\href{https://github.com/hendrycks/imagenet-r}{ImageNet-R}}~\cite{hendrycks2021many}: A variation of ImageNet designed to evaluate the susceptibility to spurious correlations of ImageNet models. It includes 30,000 artistic renditions (e.g., paintings, origami, or sculptures) of 200 ImageNet object classes. The images were primarily collected from \textit{Flickr}\footnote{\url{https://www.flickr.com/}}. $\xrightarrow[]{}$~\textbf{Used by}:~\cite{Mitrovic_2020_relic}
    
    \item \textbf{\href{https://github.com/mgbellemare/Arcade-Learning-Environment}{The Arcade Learning Environment (ALE)}}~\cite{bellemare13arcade}: A suite of Atari 2600 games that allows researchers to develop AI agents (mostly RL agents) for more than 100 games. ALE supports OpenAI gym, Python, and C++ and provides researchers with a plethora of features to evaluate different agents. $\xrightarrow[]{}$~\textbf{Used by}:~\cite{Mitrovic_2020_relic, goyal2021recurrent}
    
    \item \textbf{\href{https://www.cs.toronto.edu/~kriz/cifar.html}{CIFAR}}~\cite{krizhevsky2009learning}: The two CIFAR datasets, CIFAR-10 and CIFAR-100, are labeled images stemming from the now withdrawn Tiny Images dataset\footnote{\url{http://groups.csail.mit.edu/vision/TinyImages/}}. The more prominent set, CIFAR-10, contains 60000 32 $\times$ 32 color images separated into ten mutually exclusive classes, with 6000 images per class. CIFAR-100 is simply a 100-class version of CIFAR-10. $\xrightarrow[]{}$~\textbf{Used by}:~\cite{ilyas2022datamodels, zhang2021causaladv}
    
    \item \textbf{\href{https://github.com/fMoW/dataset}{Functional Map of the World (FMoW)}}~\cite{christie2018functional}: A collection of over 1 million satellite images depicting more than 200 countries. Each satellite contains at least one of 63 box annotations categorizing visible landmarks, such as \textit{flooded road} or \textit{airport}. $\xrightarrow[]{}$~\textbf{Used by}:~\cite{ilyas2022datamodels}
    
    \item \textbf{\href{https://github.com/dido1998/CausalMBRL}{Chemical Environment}}~\cite{ke2021systematic}: This synthetic environment was designed to evaluate causal reinforcement learning (RL) agents exhaustively. In this task, agents must change the colors of a given set of objects. However, altering one object influences the color of other objects. The causal dynamics are set by either a user-defined causal graph or a randomly generated DAG. $\xrightarrow[]{}$~\textbf{Used by}:~\cite{Wang_2022_CDL}
    
    \item \textbf{\href{https://github.com/ARISE-Initiative/robosuite}{robosuite}}~\cite{zhu2020robosuite}: A simulation framework built upon the MuJoCo physics engine allowing researchers to simulate contact dynamics for robot learning tasks. Given a set of cubes, RL agents must maneuver a robotic arm to solve different tasks (e.g., stacking the cubes or lifting one to a specified height). $\xrightarrow[]{}$~\textbf{Used by}:~\cite{Wang_2022_CDL}
    
    \item \textbf{\href{https://github.com/propublica/compas-analysis/}{COMPAS Recidivism Risk}}~\cite{angwin2016COMPAS}: A set of criminological datasets published by ProPublica to evaluate the bias of COMPAS - an algorithm used to assess the likelihood of criminal defendants reoffending. All COMPAS-related datasets include data from over 10,000 defendants, each being described via 52 features (e.g., age, sex, race) and with a label indicating whether they were rearrested within two years. $\xrightarrow[]{}$~\textbf{Used by}:~\cite{dominguez2022adversarial}
    
    \item \textbf{\href{https://archive.ics.uci.edu/ml/datasets/adult}{Adult (Census Income)}}~\cite{kohavi1996ADULTS, dheeru2017uci}: A tabular dataset containing anonymized data from the 1994 Census bureau database.\footnote{\url{http://www.census.gov/en.html}} Classifiers try to predict whether a given person will earn over or under 50,000 USD worth of salary. Each person is described via 15 features (including their id), e.g., gender, education, and occupation. $\xrightarrow[]{}$~\textbf{Used by}:~\cite{dominguez2022adversarial}
    
    \item \textbf{\href{https://archive.ics.uci.edu/ml/datasets/South+German+Credit+\%28UPDATE\%29}{South German Credit}}~\cite{groemping2019south}: Designed as a successor to the German Credit dataset, this dataset contains 1000 credit scoring entries from a south german bank between 1973 and 1975. Each row contains 20 columns (e.g., savings, job, and credit history) based on which models must assess the risk of granting credit. $\xrightarrow[]{}$~\textbf{Used by}:~\cite{dominguez2022adversarial}
    
    \item \textbf{\href{https://github.com/RicardoDominguez/AdversariallyRobustRecourse/tree/main/data}{Bail (DATA 1978)}}~\cite{schmidt1988predicting}: A collection of criminal records from 9,327 individuals that were released from a North Carolina prison between 1977 and 1978. This dataset was created to investigate factors that influence the likelihood of recidivism. Each record contains 19 variables, including a binary ethnicity variable (black or not black) and a variable indicating previous use of hard drugs. $\xrightarrow[]{}$~\textbf{Used by}:~\cite{dominguez2022adversarial}
    
    \item \textbf{\href{https://github.com/IBM/OoD/tree/master/IRM_games}{Colored FashionMNIST}}~\cite{ahuja2020invariant}: This dataset was inspired by~\citet{arjovsky2019invariant} Colored MNIST dataset.~\citeauthor{ahuja2020invariant} use the same coloring approach to induce spurious correlations into FashionMNIST data (greyscaled Zalando articles). $\xrightarrow[]{}$~\textbf{Used by}:~\cite{lu2021invariant}
    
    \item \textbf{\href{https://github.com/belaalb/G2DM\#download-vlcs}{VLCS}}~\cite{torralba2011unbiased}: A collection of 10,729 images from four standard datasets designed to evaluate the OOD performance of image classifiers. Each image depicts a bird, car, chair, dog, or person. $\xrightarrow[]{}$~\textbf{Used by}:~\cite{lu2021invariant}

    \item \textbf{\href{https://www.iro.umontreal.ca/~agrawal/vqa-cp/}{VQA-CP}}~\cite{agrawal2018don}: A dataset for Visual Question Answering (VQA) models that actively punishes the use of spurious correlations. This is achieved by rearranging the VQA v1 and VQA v2 data splits. The resulting training and test data differ in the "distribution of answers per question type". $\xrightarrow[]{}$~\textbf{Used by}:~\cite{teney2020learning}
    
    \item \textbf{\href{https://cocodataset.org/\#download}{COCO}}~\cite{lin2014microsoft}: An object detection dataset containing 328k images that depict 91 different types of objects. Each object within an image has its unique annotation, leading to more than 2.5 million labels across the entire dataset. $\xrightarrow[]{}$~\textbf{Used by}:~\cite{teney2020learning}

    \item \textbf{\href{http://www.seaphe.org/databases.php}{Law School Admission Data}}~\cite{LawSchoolAdmission}: A tabular dataset of admission data from 25 US law schools between 2005 and 2007. This dataset contains information from more than 100,000 applicants (e.g., gender, ethnic group, LSAT score), with each entry having a binary admission status variable. $\xrightarrow[]{}$~\textbf{Used by}:~\cite{wang2021enhancing}

    \item \textbf{\href{http://yann.lecun.com/exdb/mnist/}{MNIST}}~\cite{lecun1998gradient}: An extraordinarily well-known and widely used image dataset comprising 28 $\times$ 28 grayscale images of handwritten digits. It contains 60,000 training and 10,000 test samples. $\xrightarrow[]{}$~\textbf{Used by}:~\cite{zhang2021causaladv, zhang2020causal, little2019causal}
    
    \item \textbf{\href{https://github.com/teganmaharaj/zoneout}{Sequential MNIST Resolution Task}}~\cite{krueger2016zoneout}: A sequential version of MNIST, where pixels of an handwritten digit are shown one at a time.  $\xrightarrow[]{}$~\textbf{Used by}:~\cite{goyal2021recurrent}
    
    \item \textbf{\href{https://github.com/sjoerdvansteenkiste/Relational-NEM}{Bouncing Ball}}~\cite{van2018relational}: A simulation environment where multiple balls of different sizes and weights independently move according to Newtonian physics. This environment is used to assess the model's physical reasoning capabilities under different conditions (e.g., different amounts of balls). $\xrightarrow[]{}$~\textbf{Used by}:~\cite{goyal2021recurrent}
    
    \item \textbf{\href{https://github.com/mila-iqia/babyai}{BabyAI}}~\cite{babyai_iclr19}: A RL framework that supports the development of agents that can understand language instructions. For this purpose, the authors developed agents that simulate human experts capable of communicating with task-solving agents using synthetic natural language. The platform provides 19 levels to alter the difficulty of the task. $\xrightarrow[]{}$~\textbf{Used by}:~\cite{goyal2021recurrent}
    
    \item \textbf{ETH and UCY}~\cite{lerner2007crowds, ess2007eth}: Both \textit{\href{https://data.vision.ee.ethz.ch/cvl/aess/dataset/}{ETH}}~\cite{ess2007eth} and \textit{\href{https://github.com/CHENGY12/CausalHTP}{UCY}}~\cite{lerner2007crowds} are datasets containing real-world pedestrian trajectories. More novel papers combine both datasets to simulate multiple training and testing environments. Together, they contain trajectories of 1536 detected pedestrians collected from five locations. $\xrightarrow[]{}$~\textbf{Used by}:~\cite{liu2022towards, Chen_2021_posthoc}
    
    \item \textbf{\href{https://cvgl.stanford.edu/projects/uav_data/}{Stanford Drone dataset}}~\cite{robicquet2016learning}: A video dataset containing over 100 top-view scenes of the Stanford University campus that were shot with a quadcopter. The videos depict 20,000 manually annotated targets (e.g., pedestrians, bicyclists, or cars). $\xrightarrow[]{}$~\textbf{Used by}:~\cite{Chen_2021_posthoc, liu2022towards}
    
    \item \textbf{\href{https://github.com/kohpangwei/group_DRO}{Waterbirds}}~\cite{sagawa2019distributionally}: A binary image classification task where models must decide whether the depicted bird is a waterbird or a landbird. Good-performing models must rely on something other than the intrinsic spurious correlation between the background and the label (e.g., only 56 out of 4795 training images depict a waterbird with a land background). $\xrightarrow[]{}$~\textbf{Used by}:~\cite{wang2022ISR}
    
    \item \textbf{\href{https://mmlab.ie.cuhk.edu.hk/projects/CelebA.html}{CelebA}}~\cite{liu2015faceattributes}: A face image dataset containing 202,599 images of size 178×218 from 10,177 unique celebrities. Each image is annotated with 40 binary facial attributes (e.g., \textit{Is this person smiling?}) and five landmark positions describing the 2D position of the eyes, the nose, and the mouth (split into \textit{left} and \textit{right} side of the mouth). $\xrightarrow[]{}$~\textbf{Used by}:~\cite{wang2022ISR}
    
    \item \textbf{\href{https://cims.nyu.edu/~sbowman/multinli/}{MultiNLI}}~\cite{williams2017broad}: A text dataset developed to evaluate natural language inference (NLI) models. Models must decide whether a given hypothesis is contradictory to, entailed by, or neutral to the given premise. Contrary to other NLI datasets, MultiNLI includes text from 10 written and spoken English domains. $\xrightarrow[]{}$~\textbf{Used by}:~\cite{wang2022ISR}

    \item \textbf{\href{https://archive.ics.uci.edu/ml/datasets/abalone}{Abalone}}~\cite{dheeru2017uci}: In this task, ML models need to predict the number of rings an \textit{abalone} (a shellfish) has based on the given features \textit{sex}, \textit{width}, \textit{height}, and \textit{shell diameter}. The dataset contains 4177 entries. $\xrightarrow[]{}$~\textbf{Used by}:~\cite{kyono2019improving}

    \item \textbf{\href{https://www.kaggle.com/datasets/marklvl/bike-sharing-dataset}{Bike Sharing in Washington D.C.}}~\cite{fanaee2014event}: This dataset contains the hourly and daily count of rental bikes used in Washington D.C. between 2011 and 2012 (17,379 entries). Given weather and seasonal information, models need to predict the count of total rental bikes. $\xrightarrow[]{}$~\textbf{Used by}:~\cite{kyono2019improving}

    \item \textbf{\href{https://www.kaggle.com/datasets/open-powerlifting/powerlifting-database}{OpenPowerlifting}}~\cite{OpenPowerlifting2019}: This powerlifting competition dataset includes more than 22,000 competitions and more than 412,000 competitors as of April 2019. The data stem from OpenPowerlifting\footnote{\url{https://www.openpowerlifting.org/}}, with each entry containing information about the lifter, the equipment used, weight class, and their performance across different powerlifting disciplines.  $\xrightarrow[]{}$~\textbf{Used by}:~\cite{kyono2019improving}
    
\end{itemize}

\subsection{Interesting Causal Tools}

\begin{itemize}

    \item \textbf{\href{https://github.com/causal-disentanglement/CANDLE}{CANDLE}}~\citep{reddy:2022:AAAI:candle_disentangled_dataset}: A dataset of realistic images of objects in a specific scene generated based on observed and unobserved confounders (object, size, color, rotation, light, and scene). As each of the 12546 images is annotated with the ground-truth information of the six generating factors, it is possible to emulate interventions on image features.

    \item \textbf{\href{https://github.com/rr-learning/CausalWorld}{CausalWorld}}~\cite{ahmed2021causalworld}: A simulation framework and benchmark that provides RL agents different learning tasks in a robotic manipulation environment. The environment comes with a causal structure on which users and agents can intervene on variables such as object masses, colors or sizes.
    
    \item \textbf{\href{https://github.com/huawei-noah/trustworthyAI/tree/master/gcastle}{gCastle}}~\cite{zhang2021gcastle}: An end-to-end causal structure learning toolbox that is equipped with 19 techniques for Causal Discovery. It also assists users in data generation and evaluating learned structures. Having a firm understanding of the causal structure allows models to deduce the content and style variables of the domain.
    
    \item \textbf{\href{https://github.com/felixleopoldo/benchpress}{Benchpress}}~\cite{rios2021benchpress}: A benchmark for causal structure learning allowing users to compare their causal discovery methods with over 40 variations of state-of-the-art algorithms. The plethora of available techniques in this single tool could facilitate research into robustness of ML systems through causality.
    
\end{itemize}

\subsection{Prominent Non-Causal Tools}

\begin{itemize}

    \item \textbf{DomainBed and OOD-Bench}~\cite{Gulrajani_2020_DomainBed, Ye2021ood}: \href{https://github.com/facebookresearch/DomainBed}{DomainBed} is a benchmark for OOD-learning that enables performance comparisons with more than 20 OOD-algorithms on 10 different, popular OOD-datasets. \href{https://github.com/m-Just/OoD-Bench}{OOD-Bench} is built upon DomainBed and introduces a measurement to quantify the Diversity shift and Correlation shift inherit to OOD-datasets. The resulting categorization allows researchers to pinpoint strengths and weaknesses of OOD-learning algorithms.
    
    \item \textbf{\href{https://github.com/RobustBench/robustbench}{RobustBench}}~\cite{croce2020robustbench}: A standardized adversarial robustness benchmark capable of emulating a variety of adversarial attacks for image classification through \textit{AutoAttack}. It also provides multiple continuously updated leaderboards of the most robust models, which allows for direct comparisons between causal and non-causal methods. 
    
    \item \textbf{\href{https://github.com/bethgelab/foolbox}{Foolbox}}~\cite{rauber2017foolboxnative}: A popular Python library that allows researchers to test their adversarial defenses against state-of-the-art adversarial attacks. Foolbox is very compatible, natively supporting Pytorch, Tensorflow and JAX models.
    
    \item \textbf{\href{https://github.com/AI-secure/VeriGauge}{VeriGauge}}~\cite{Li_2020_SoK}: A Python toolbox that allows users to verify the robustness of their adversarial defense approach for deep neural networks. It not only covers a multitude of verification techniques but also comes with an up-to-date leaderboard.
    
    \item \textbf{\href{https://github.com/Trusted-AI/adversarial-robustness-toolbox}{Adversarial Robustness Toolbox (ART)}}~\cite{art2018}: An extensive Python library for Adversarial Machine Learning. It not only equips researchers with various attacks and defenses across four different attack threats (evasion, extraction, poisoning, and inference) but also provides the means to assess the performance of such algorithms thoroughly. ART is compatible with many popular frameworks and supports various data types and learning tasks.
    
\end{itemize}

\section{Privacy}
\label{sec:appendix_privacy}

\subsection{Datasets Used by Cited Publications}

\begin{itemize}

    \item \textbf{\href{https://github.com/ghif/mtae}{Rotated MNIST}}~\cite{ghifary2015domain}: The dataset consists of MNIST images with each domain containing images rotated by a particular angle $\{0^{\circ}, 15^{\circ}, 30^{\circ}, 45^{\circ}, 60^{\circ}, 75^{\circ}\}$ $\xrightarrow[]{}$~\textbf{Used by}:~\cite{de2022mitigating, francis2021towards}
    
    \item \textbf{\href{https://github.com/facebookresearch/DomainBed}{PACS}}~\cite{li2017deeper}: An image classification dataset categorized into 10 classes that are scattered across four different domains, each having a distinct trait: photograph, art, cartoon and sketch. $\xrightarrow[]{}$~\textbf{Used by}:~\cite{de2022mitigating}
    
    \item \textbf{\href{https://www.hemanthdv.org/officeHomeDataset.html}{Office-Home}}~\cite{venkateswara2017deep}: Image classification dataset analogous to PACS, having four distinct image domains: Art, ClipArt, Product and Real-World. $\xrightarrow[]{}$~\textbf{Used by}:~\cite{de2022mitigating}
   
    \item \textbf{\href{https://github.com/facebookresearch/InvariantRiskMinimization}{ColoredMNIST}}~\cite{arjovsky2019invariant}: The dataset consists of input images with digits 0-4 colored red and labelled 0 while digits 5-9 are colored green representing the two domains. $\xrightarrow[]{}$~\textbf{Used by}:~\cite{francis2021towards, gupta2022fl}
    
    \item \textbf{\href{https://github.com/IBM/OoD/tree/master/IRM_games}{Colored FashionMNIST}}~\cite{ahuja2020invariant}: This dataset was inspired by~\citet{arjovsky2019invariant} Colored MNIST dataset.~\citeauthor{ahuja2020invariant} use the same coloring approach to induce spurious correlations into FashionMNIST data (greyscaled Zalando articles). $\xrightarrow[]{}$~\textbf{Used by}:~\cite{gupta2022fl}

    \item \textbf{\href{https://www.cs.toronto.edu/~kriz/cifar.html}{CIFAR}}~\cite{krizhevsky2009learning}: The two CIFAR datasets, CIFAR-10 and CIFAR-100, are labeled images stemming from the now withdrawn Tiny Images dataset\footnote{\url{http://groups.csail.mit.edu/vision/TinyImages/}}. The more prominent set, CIFAR-10, contains 60000 32 $\times$ 32 color images separated into ten mutually exclusive classes, with 6000 images per class. CIFAR-100 is simply a 100-class version of CIFAR-10. $\xrightarrow[]{}$~\textbf{Used by}:~\cite{gupta2022fl,jiang2021tsmobn}
    
    \item \textbf{\href{https://github.com/KaiyangZhou/Dassl.pytorch/blob/master/DATASETS.md\#digits-dg}{Digits-DG}}~\cite{zhou2020learning}: An image dataset specifically designed to evaluate the performance of models on OOD data. It includes images from four different handwritten digits databases. Each dataset represents a unique domain as images from different datasets significantly differ in terms of, e.g., handwriting style or background color. $\xrightarrow[]{}$~\textbf{Used by}:~\cite{jiang2021tsmobn}
    
    \item \textbf{\href{https://camelyon17.grand-challenge.org/Data/}{Camelyon17}}~\cite{bandi2018detection}: A publicly available medical dataset containing 1000 histology images from five Dutch hospitals. Given an image, classification models need to detect breast cancer metastases. $\xrightarrow[]{}$~\textbf{Used by}:~\cite{jiang2021tsmobn}
   
\end{itemize}

\subsection{Interesting Causal Tools For Federated Learning}

The publications reviewed in \cref{sec:privacy} are largely causal approaches to Federated Learning (FL). As such, we mainly provide an overview of causal and non-causal tools for FL.

\begin{itemize}

    \item \textbf{Federated Causal Discovery}~\cite{abyaneh2022fed, gao2021federated}: Until this point, we suggested general causal discovery tools like \textit{gCastle}~\cite{zhang2021gcastle} or \textit{benchpress}~\cite{rios2021benchpress}. However, the provided methods translate poorly into the federated setting due to the decentralized data. As such, we would like to refer readers to recently developed \textbf{Federated Causal Discovery} techniques (e.g.,~\cite{abyaneh2022fed, gao2021federated}). These methods are specifically designed to conduct causal discovery on decentralized data in a privacy-preserving manner.

    \item \textbf{\href{https://github.com/causal-disentanglement/CANDLE}{CANDLE}}~\cite{reddy:2022:AAAI:candle_disentangled_dataset}: A dataset of realistic images of objects in a specific scene generated based on observed and unobserved confounders (object, size, color, rotation, light, and scene). As each of the 12546 images is annotated with the ground-truth information of the six generating factors, it is possible to emulate interventions on image features. Users/Devices could be simulated by altering the scenery.
    
    \item \textbf{\href{https://github.com/vothanhvinh/FedCI}{Federated Causal Effect Estimation}}~\cite{vo2022bayesian}: Similar to causal discovery, standard causal effect estimation methods were not designed for decentralized data. Only very recently,~\citeauthor{vo2022bayesian} developed a causal effect estimation framework compatible with federated learning. Despite this line of work's infancy, we believe that this publication is important for more privacy-preserving causal learning.
    
\end{itemize}

\subsection{Prominent Non-Causal Federated Learning Tools}

\begin{itemize}

    \item \textbf{\href{https://github.com/TalwalkarLab/leaf}{LEAF}}~\cite{caldas2018leaf}: A benchmark containing datasets explicitly designed to analyze FL algorithms. The six datasets include existing re-designed databases such as \textit{CelebA}~\cite{liu2015faceattributes} to emulate different devices/users and newly created datasets. LEAF also provides evaluation methods and baseline reference implementations for each dataset.
    
    \item \textbf{\href{https://github.com/Di-Chai/FedEval}{FedEval}}~\cite{chai2020fedeval}: A publicly available evaluation platform for FL. It allows researchers to compare their FL methods with existing state-of-the-art algorithms on seven datasets based on five FL-relevant metrics (Accuracy, Communication, Time efficiency, Privacy, and Robustness). The benchmark utilizes Docker container technology to simulate the server and clients and socket IO for simulating communication between the two.
    
    \item \textbf{\href{https://github.com/Xtra-Computing/OARF}{OARF}}~\cite{hu2022oarf}: An extensive benchmark suite designed to assess state-of-the-art FL algorithms for both horizontal and vertical FL. It includes 22 datasets that cover different domains for both FL variants. Additionally, OARF provides several metrics to evaluate FL algorithms, and its modular design enables researchers to test their own methods.
    
    \item \textbf{\href{https://github.com/FedML-AI/FedML/tree/master/python/app/fedgraphnn}{FedGraphNN}}~\cite{he2021fedgraphnn}: An FL benchmark for Graph Neural Networks (GNN). In order to provide a unified platform for the development of graph-based FL solutions, FedGraphNN supplies users with 36 graph datasets across seven different domains. Researchers can also employ and compare their own \textit{PyTorch (Geometric)} models with different GNNs.

    \item \textbf{\href{https://github.com/liuyugeng/ML-Doctor}{ML-Doctor}}~\cite{mldoctor_2022}: A codebase initially used to compare and evaluate different inference attacks (membership inference, model stealing, model inversion, and attribute inference). Its modular structure enables researchers to assess the effectiveness of their privacy-preserving algorithms against SOTA privacy attacks. 
    
\end{itemize}

\section{Safety}
\label{sec:appendix_safety}

\subsection{Datasets Used by Cited Publications}

\begin{itemize}

    \item \textbf{\href{https://www.sciencedirect.com/}{ScienceDirect}}~\cite{ScienceDirect_2023}: A bibliographic database that hosts over 18 million publications from more than 4,000 journals and more than 30,000 e-books from the publisher Elsevier. Launched back in 1997, ScienceDirect includes papers from engineering and medical research areas and social sciences and humanities.$\xrightarrow[]{}$~\textbf{Used by}:~\cite{voegeli2019sustainability}
    
    \item \textbf{\href{https://data.worldbank.org/indicator}{World Bank}}~\cite{WDI}: A publicly available collection of datasets that facilitate the analysis of global development. Researchers can use this data to compare countries under different developmental aspects, including agricultural progress, poverty, population dynamics, and economic growth. $\xrightarrow[]{}$~\textbf{Used by}:~\cite{wu2015causality}

    \item \textbf{World Economic Forum (WEF)}~\cite{WEF_2023}: The WEF is an international non-governmental based in Switzerland that publishes economic reports such as the Global Competitiveness Report. The reports are available online, with some of the data being easily accessible through websites like \href{https://knoema.com/atlas/sources/WEF}{Knoema}. $\xrightarrow[]{}$~\textbf{Used by}:~\cite{haseeb2019economic}

    \item \textbf{\href{https://stats.oecd.org/}{OECD.Stat}}~\cite{OECD_stats_2023}: This webpage includes data and metadata for OECD countries and selected non-member economies. The online platform allows researchers to traverse the collected data through given data themes or via search-engine queries. $\xrightarrow[]{}$~\textbf{Used by}:~\cite{haseeb2019economic}

    \item \textbf{\href{https://branddb.wipo.int/en/}{Global Brand Database}}~\cite{WIPO_2023}: An online database hosted by the World Intellectual Property Organization (WIPO) that contains information about Trademark applications (e.g., owner of the trademark, its status, or the designation country). It currently contains almost 53 million records from 73 data sources. $\xrightarrow[]{}$~\textbf{Used by}:~\cite{haseeb2019economic}
   
    \item \textbf{\href{https://pubmed.ncbi.nlm.nih.gov/}{PubMed}}~\cite{PubMed2022}: A widely-known, free-to-access search engine for biomedical and life science literature developed and maintained by the National Center for Biotechnology Information (NCBI). Researchers can find more than 34 million citations and abstracts of articles. PubMed does not host the articles themselves but frequently provides a link to the full-text articles.  $\xrightarrow[]{}$~\textbf{Used by}:~\cite{gillespie2021impact}
   
    \item \textbf{\href{https://www.proquest.com/}{ProQuest Central}}~\cite{ProQuest2022}: A database containing dissertations and theses in a multitude of disciplines. It currently contains more than 5 million graduate works. $\xrightarrow[]{}$~\textbf{Used by}:~\cite{gillespie2021impact}
   
    \item \textbf{\href{https://www.cochranelibrary.com/central}{Cochrane Central Register of Controlled Trials (CENTRAL)}}~\cite{Cochrane2022}: A database of reports for randomized and quasi-randomized controlled trials collected from different online databases. Although it does not contain full-text articles, the CENTRAL includes bibliographic details and often an abstract of the report. $\xrightarrow[]{}$~\textbf{Used by}:~\cite{gillespie2021impact}
   
    \item \textbf{\href{https://www.apa.org/pubs/databases/psycinfo/index}{PsycINFO}}~\cite{PsycINFO2022}: A database hosted and developed by American Psychological Association containing abstracts for more than five million articles in the field of psychology. $\xrightarrow[]{}$~\textbf{Used by}:~\cite{gillespie2021impact}

    \item \textbf{\href{https://www.kaggle.com/datasets/wordsforthewise/lending-club}{Lending Club}}~\cite{Lending_Club_2018}: A dataset that contains information about all accepted and rejected peer-to-peer loan applications of LendingClub. Currently, the data are only available through the referenced Kaggle entry, as the company no longer provides peer-to-peer loan services\footnote{\url{https://www.lendingclub.com/investing/peer-to-peer}}. $\xrightarrow[]{}$~\textbf{Used by}:~\cite{tsirtsis2020decisions}

    \item \textbf{\href{https://github.com/ustunb/actionable-recourse/tree/master/examples/paper/data}{Taiwanese Credit Data}}~\cite{ustun2019actionable, yeh2009comparisons}: A real-world dataset containing payment data collected in October 2005 from a Taiwanese bank. The commonly used pre-processed version\footnote{Available at \url{https://github.com/ustunb/actionable-recourse/tree/master/examples/paper/data} under the name "credit\_processed.csv"} \cite{ustun2019actionable} contains data from 30,000 individuals described through 16 features (e.g., marital status, age, or payment history). $\xrightarrow[]{}$~\textbf{Used by}:~\cite{tsirtsis2020decisions}

\end{itemize}

\subsection{Interesting Causal Tools}

\begin{itemize}

    \item \textbf{\href{https://google.github.io/CausalImpact/CausalImpact.html}{CausalImpact}}~\cite{brodersen2015inferring}: This R package allows users to conduct causal impact assessment for planned interventions on serial data given a response time series and an assortment of control time series. For this purpose, CausalImpact enables the construction of a Bayesian structural time-series model that can be used to predict the resulting counterfactual of an intervention.
    
    \item \textbf{\href{https://github.com/BiomedSciAI/causallib}{Causal Inference 360}}~\cite{shimoni2019evaluation}: A Python package developed by IBM to infer causal effects from given data. Causal Inference 360 includes multiple estimation methods, a medical dataset, and multiple simulation sets. The provided methods can be used for any complex ML model through a scikit-learn-inspired API.
    
    \item \textbf{\href{https://github.com/huawei-noah/trustworthyAI/tree/master/gcastle}{gCastle}}~\cite{zhang2021gcastle}: An end-to-end causal structure learning toolbox that is equipped with 19 techniques for Causal Discovery. It also assists users in data generation and evaluating learned structures. Having a firm understanding of the causal structure is crucial for safety-related research.
    
    \item \textbf{\href{https://github.com/felixleopoldo/benchpress}{Benchpress}}~\cite{rios2021benchpress}: A benchmark for causal structure learning allowing users to compare their causal discovery methods with over 40 variations of state-of-the-art algorithms. The plethora of available techniques in this single tool could facilitate research into safety and accountability of ML systems through causality.
    
    \item \textbf{\href{https://webdav.tuebingen.mpg.de/cause-effect/}{CauseEffectPairs}}~\cite{mooij2016distinguishing}: A collection of more than 100 databases, each annotated with a two-variable cause-effect relationship (e.g., access to drinking water affects infant mortality). Given a database, models need to distinguish between the cause and effect variables.
    
\end{itemize}

\subsection{Prominent Non-Causal Tools}

\begin{itemize}
    
    \item \textbf{\href{https://github.com/canada-ca/aia-eia-js}{Government of Canada's AIA tool}}~\cite{CanadaAIA}: The Algorithmic Impact Assessment (AIA) tool is a questionnaire developed in the wake of Canada's Directive on Automated Decision Making\footnote{\url{http://www.tbs-sct.gc.ca/pol/doc-eng.aspx?id=32592}}. Employees of the Canadian Government wishing to employ automatic decision-making systems in their projects first need to assess the impact of such systems via this tool. Based on answers given to ca. 80 questions revolving around different aspects of the projects, AIA will output two scores: one indicating the risks that automation would bring and one that quantifies the quality of the risk management.
    
    \item \textbf{\href{https://github.com/dssg/aequitas}{Aequitas}}~\cite{2018aequitas}: An open-source auditing tool designed to assess the bias of algorithmic decision-making systems. It provides utility for evaluating the bias of decision-making outcomes and enables users to assess the bias of actions taken directly. 
    
    \item \textbf{\href{https://github.com/microsoft/responsible-ai-toolbox/blob/main/docs/erroranalysis-dashboard-README.md}{Error Analysis (\textit{Responsible AI})}}~\cite{ResponsibleAI}: As part of the Responsible AI toolbox, Error Analysis is a model assessment tool capable of identifying subsets of data in which the model performs poorly (e.g., black citizens being more frequently misclassified as potential re-offenders). It also enables users to diagnose the root cause of such poor performance. 

    \item \textbf{\href{https://github.com/liuyugeng/ML-Doctor}{ML-Doctor}}~\cite{mldoctor_2022}: A codebase initially used to compare and evaluate different inference attacks (membership inference, model stealing, model inversion, and attribute inference). Due to its modular structure, it can also be used as a Risk Assessment tool for analyzing the susceptibility against SOTA privacy attacks.
    
\end{itemize}

\section{Healthcare}
\label{sec:appendix_healthcare}

\subsection{Datasets Used by Cited Publications}

\begin{itemize}

    \item \textbf{\href{https://adni.loni.usc.edu/data-samples/access-data/}{Alzheimer’s Disease Neuroimaging Initiative (ADNI)}}~\cite{petersen2010alzheimer}: A medical dataset containing multi-modal information of over 5,000 volunteering subjects. ADNI includes clinical and genomic data, biospecimens, MRI, and PET images. Researchers need to apply for data access. $\xrightarrow[]{}$~\textbf{Used by}:~\cite{Sanchez2022, shen2020}

    \item \textbf{\href{https://www.ncbi.nlm.nih.gov/geo/query/acc.cgi?acc=GSE147507}{SARS-CoV-2 infected cells (Series GSE147507)}}~\cite{blanco2020imbalanced}: A genomic dataset that contains RNA-seq data from SARS-CoV-2-infected cells of both humans and ferrets. The data are publicly available on the NCBI Gene Expression Omnibus (GEO) server under the accession number GSE147507. $\xrightarrow[]{}$~\textbf{Used by}:~\cite{belyaeva2021}

    \item \textbf{\href{https://gtexportal.org/home/}{The Genotype-Tissue Expression (GTEx) project}}~\cite{carithers2015novel}: An online medical platform that provides researchers with tissue data. Data samples stem from 54 non-diseased tissue sites across nearly 1000 individuals whose genomes were processed via sequencing methods such as WGS, WES, and RNA-Seq. $\xrightarrow[]{}$~\textbf{Used by}:~\cite{belyaeva2021}

    \item \textbf{\href{https://www.ncbi.nlm.nih.gov/geo/query/acc.cgi?acc=GSE92742}{L1000 Connectivity Map (Series GSE92742)}}~\cite{subramanian2017next}: A connectivity map (CMap) connects genes, drugs, and disease states based on their gene-expression signatures. The CMap provided by~\citeauthor{subramanian2017next} includes over 1.3 million L1000 profiles of 25,200 unique perturbagens~\footnote{See \url{https://clue.io/connectopedia/perturbagen_types_and_controls} for the definition of this term}. The data are publicly available on the NCBI Gene Expression Omnibus (GEO) server under the accession number GSE92742.$\xrightarrow[]{}$~\textbf{Used by}:~\cite{belyaeva2021}

    \item \textbf{\href{https://irefindex.vib.be/wiki/index.php/iRefIndex}{iRefIndex}}~\cite{razick2008irefindex}: This protein-protein interaction (PPI) network is a graph-based database of molecular interactions between proteins from over ten organisms. The current version of iRefIndex (version 19) contains over 1.6 million PPIs. $\xrightarrow[]{}$~\textbf{Used by}:~\cite{belyaeva2021}

    \item \textbf{\href{https://drugcentral.org/}{DrugCentral}}~\cite{avram2021drugcentral}: An online platform that provides up-to-date drug information. Users can traverse the database online through the corresponding website or via an API. The platform currently contains information on almost 5,000 active ingredients. $\xrightarrow[]{}$~\textbf{Used by}:~\cite{belyaeva2021}

    \item \textbf{\href{https://www.ncbi.nlm.nih.gov/geo/query/acc.cgi?acc=GSE81861}{Colorectal Cancer Single-cell Data (GSE81861)}}~\cite{li2017reference}: The authors provide two datasets. The first dataset contains 1,591 single cells RNA-seq data from 11 colorectal cancer patients. The second dataset contains 630 single cells from seven cell lines and can be used to benchmark cell-type identification algorithms. The data are publicly available on the NCBI Gene Expression Omnibus (GEO) server under the accession number GSE81861. $\xrightarrow[]{}$~\textbf{Used by}:~\cite{belyaeva2021}

    \item \textbf{\href{https://www.nupulmonary.org/resources/}{Pulmonary Fibrosis Single-cell Data}}~\cite{reyfman2019single}: This genomic dataset contains approximately 76,000 single-cell RNA-seq data from healthy lungs and lungs from patients with pulmonary fibrosis. The data are available online and comes with a cluster visualization based on marker gene expressions. $\xrightarrow[]{}$~\textbf{Used by}:~\cite{belyaeva2021}

    \item \textbf{\href{https://www.ndexbio.org/\#/network/5d97a04a-6fab-11ea-bfdc-0ac135e8bacf}{SARS-CoV-2 Host-Pathogen Interaction Map}}~\cite{gordon2020sars}: A PPI network that maps 27 SARS-CoV-2 proteins to human proteins through 332 high-confidence protein-protein interactions. The online data contain data from the initial study and the CORUM database \cite{tsitsiridis2023corum}. $\xrightarrow[]{}$~\textbf{Used by}:~\cite{belyaeva2021}

    \item \textbf{\href{https://wiki.cancerimagingarchive.net/pages/viewpage.action?pageId=1966254}{Lung Image Database Consortium image collection (LIDC-IDRI)}}~\cite{armato2011lung}: An image dataset comprising annotated thoracic CT Scans of more than 1,000 cases. The data stem from seven academic centers and eight medical imagining companies. Four trained thoracic radiologists provided the image annotations. $\xrightarrow[]{}$~\textbf{Used by}:~\cite{vanAmsterdam2019}

    \item \textbf{\href{https://www.nlm.nih.gov/medline/medline_overview.html}{MEDLINE}}~\cite{medline2023}: An online bibliographic database of more than 29 million article references from the field of life science (primarily in biomedicine). MEDLINE is a primary component of PubMed and is hosted and managed by the NLM National Center for Biotechnology Information (NCBI). $\xrightarrow[]{}$~\textbf{Used by}:~\cite{ziff2015}

    \item \textbf{\href{https://www.cochranelibrary.com/central}{Cochrane Central Register of Controlled Trials (CENTRAL)}}~\cite{Cochrane2022}: A database of reports for randomized and quasi-randomized controlled trials collected from different online databases. Although it does not contain full-text articles, the CENTRAL includes bibliographic details and often an abstract of the report. $\xrightarrow[]{}$~\textbf{Used by}:~\cite{ziff2015}
    
\end{itemize}

\subsection{Interesting Causal Tools}

\begin{itemize}

    \item \textbf{\href{https://github.com/BiomedSciAI/causallib}{Causal Inference 360}}~\cite{shimoni2019evaluation}: A Python package developed by IBM to infer causal effects from given data. Causal Inference 360 includes multiple estimation methods, a medical dataset, and multiple simulation sets. The provided methods can be used for any complex ML model through a scikit-learn-inspired API.
    
    \item \textbf{\href{https://github.com/huawei-noah/trustworthyAI/tree/master/gcastle}{gCastle}}~\cite{zhang2021gcastle}: An end-to-end causal structure learning toolbox that is equipped with 19 techniques for Causal Discovery. It also assists users in data generation and evaluating learned structures. Having a firm understanding of the causal structure is crucial for healthcare-related research.
    
    \item \textbf{\href{https://github.com/felixleopoldo/benchpress}{Benchpress}}~\cite{rios2021benchpress}: A benchmark for causal structure learning allowing users to compare their causal discovery methods with over 40 variations of state-of-the-art algorithms. The plethora of available techniques in this single tool could encourage more causality-based solutions for the healthcare domain.

    \item \textbf{\href{https://github.com/uber/causalml}{CausalML}}~\cite{Chen2020_CausalML_Uber}: The Python package enables users to analyze the Conditional Average Treatment Effect (CATE) or Individual Treatment Effect (ITE) observable in experimental data. The package includes tree-based algorithms, meta-learner algorithms, instrumental variable algorithms, and neural-network-based algorithms.

    \item \textbf{\href{https://github.com/mrtzh/whynot}{WhyNot}}~\cite{miller2020whynot}: A Python package that provides researchers with many simulation environments for analyzing causal inference and decision-making in a dynamic setting. It allows benchmarking of multiple decision-making systems on 13 different simulators. This set of simulators includes environments that simulate HIV treatment effects and system dynamics models of both the Zika epidemic and the US opioid epidemic.
    
\end{itemize}

\subsection{Prominent Non-Causal Tools}

\begin{itemize}
    
    \item \textbf{\href{https://github.com/Project-MONAI/MONAI}{Medical Open Network for AI (MONAI)}}~\cite{Cardoso_MONAI_An_open-source_2022}: A PyTorch-based framework that offers researchers pre-processing methods for medical imaging data, domain-specific implementations of machine learning architectures, and ready-to-use workflows for healthcare imaging. The actively maintained framework also provides APIs for integration into existing workflows.

    \item \textbf{\href{https://github.com/deepchem/deepchem}{DeepChem}}~\cite{Ramsundar-et-al-2019}: A Life Science toolbox that provides researchers with deep learning solutions for different fields of Life Science, such as Quantum Chemistry, Biology, or Drug Discovery (with the latter being particularly interesting for comparisons to causality-based solutions). Deepchem supports TensorFlow, PyTorch, and JAX and has an extensive collection of running examples.

    \item \textbf{Curated Lists on Github}~\cite{Awesome_Medical_List_2016, Medical_Data_List_2016}: 
    \citet{Awesome_Medical_List_2016} host an \href{https://github.com/kakoni/awesome-healthcare}{up-to-date GitHub repository} of relevant open-source healthcare tools and resources. \citet{Medical_Data_List_2016} provide an extensive \href{https://github.com/beamandrew/medical-data}{overview of valuable medical datasets} that could be used to assess Causal ML healthcare solutions. Although this list has not been updated since 2020, we still believe it to be a helpful initial overview of relevant datasets.
    
\end{itemize}

\end{document}